%% file: main.tex
\documentclass{article} % For LaTeX2e
\usepackage{iclr2027_conference,times}

\usepackage{hyperref}
\usepackage{url}
\usepackage{natbib}
\usepackage{algorithm}
\usepackage{algpseudocodex}
\usepackage{booktabs}
\usepackage[acronym]{glossaries}
\usepackage{graphicx}
\usepackage{enumitem}
\usepackage{xcolor}
\usepackage{siunitx}         % <-- main results table decimal right-align
\usepackage[utf8]{inputenc}  % <-- for citation names' special characters
\usepackage{multirow}        % <-- for ablation table multirows

\input{macros/math_commands}

\input{macros/gloss}

\title{Think Fast, Plan Selectively: Adaptive Deliberation for Efficient Data-Driven MPC}

\author{
    Yi~Xian~Goh\textsuperscript{\rm 1}  \qquad
    Sze~Jue~Yang\textsuperscript{\rm 2}  \qquad
    Hao~Luan\textsuperscript{\rm 3} \\
    \textsuperscript{\rm 1}Independent Researcher\\
    \textsuperscript{\rm 2}Faculty of Computer Science and Information Technology, Universiti Malaya\\
    \textsuperscript{\rm 3}School of Computing, National University of Singapore\\
    \texttt{gohyixian456@gmail.com, jasonyang429@um.edu.my,} \\
    \texttt{haoluan@comp.nus.edu.sg}
}

\newcommand{\appdxref}[1]{Appendix\textbf{~\ref{#1}}}

\iclrfinalcopy % Uncomment for camera-ready version, but NOT for submission.
\renewcommand{\headrulewidth}{0pt}
\begin{document}

\maketitle

\begin{abstract}

Data-driven model predictive control (MPC) combines learned world models with online trajectory optimization, achieving strong performance in continuous control. However, the per-step cost of sampling and evaluating hundreds of candidate trajectories restricts deployment to control frequencies well below what real-time robotics demands. Motivated by the dual-process theory of human cognition, which distinguishes between fast, intuitive processing (\emph{System~1}) and slower, deliberative reasoning (\emph{System~2}), we ask whether every decision requires the same degree of computational deliberation. We propose \textbf{Fast-TD-MPC}, a lightweight framework that adaptively routes between fast policy execution and test-time planning, reserving costly deliberation for states where it is most needed. 
Fast-TD-MPC delivers competitive task performance across 103 continuous control tasks while achieving up to ${\sim}4\times$ faster inference.
Under external disturbances, Fast-TD-MPC selectively falls back to planning, maintaining robustness comparable to the original planner.
\end{abstract}

%%Sections
%% ==================================== %%
\input{sections/intro}
\input{sections/method}

\input{sections/experiments}

\input{sections/limitations}

\input{sections/related_work}

%% ==================================== %%

%% Sections
\input{sections/conclusion}

%% ============================================================= %%
%% Stuff below this line does not count towards the 9 page limit %%
%% ============================================================= %%

\subsection*{Reproducibility statement}
Details of hyperparameters are described in the appendix.

\bibliographystyle{iclr2027_conference}
\bibliography{refs}

%%%%%%%%%% COMMENT OUT THIS at SUBMISSION %%%%%%%%%%%
\clearpage %%%% DO NOT USE THIS FOR FINAL VERSION
\appendix
\input{appendix/appendix}

%%%%%%%%%%%%%%%%%%%%%%%%%%%%%%%%%%%

\end{document}

%% file: macros/math_commands.tex
\usepackage{amsmath,amsfonts,bm}

\def\eqref#1{equation~\ref{#1}}
\def\1{\bm{1}}

\DeclareMathAlphabet{\mathsfit}{\encodingdefault}{\sfdefault}{m}{sl}
\SetMathAlphabet{\mathsfit}{bold}{\encodingdefault}{\sfdefault}{bx}{n}

%% file: macros/gloss.tex
\setacronymstyle{long-short}
\makenoidxglossaries

\newacronym{mbrl}{MBRL}{Model-based reinforcement learning}
\newacronym{mppi}{MPPI}{Model Predictive Path Integral}
\newacronym{mdp}{MDP}{Markov Decision Process}
\newacronym{td}{TD}{Temporal Difference}
\newacronym{mpc}{MPC}{Model Predictive Control}

%% file: sections/intro.tex
\section{Introduction}
\label{sec:intro}

Data-driven \gls{mpc} combines learned world models with online trajectory optimization, achieving state-of-the-art performance across diverse domains \citep{hansen2024tdmpc2,hafner2023dreamerv3,schrittwieser2020muzero}.
By learning to predict environment dynamics, world models \citep{ha2018worldmodels} unlock \emph{test-time planning}: the ability to propose, evaluate, and optimize candidate action sequences before execution, rather than reactively mapping states to actions \citep{hansen2022tdmpc,schrittwieser2020muzero,wang2025bmpc}.
This planning capability consistently outperforms purely reactive policies and enhances robustness in dynamic environments, driving progress in robotic manipulation and locomotion \citep{hansen2024tdmpc2,hafner2023dreamerv3,hafner2020dreamer,nagabandi2018neural_dynamics,hafner2019planet}, autonomous driving \citep{guan2024driving_survey,liu2026lidarad}, and game-playing \citep{schrittwieser2020muzero}.

Planning brings robustness, but at the cost of substantial per-step computation. At each decision step, a planner samples and evaluates hundreds of candidate trajectories, resulting in per-step latencies that exceed the control frequencies required for real-time robotic deployment and latency-sensitive tasks \citep{dong2024realtime_mpc,gangapurwala2022locomotion}.
When planning is slow relative to changes in the environment, the state for which an action was planned may have already changed by the time it is executed, resulting in \emph{stale-action} errors \citep{cheng2026adarep,lin2025speculation}.

This raises a natural question: \emph{does every decision require the same amount of deliberation?}

In most control problems, the answer is unlikely to be yes \citep{kahneman2011thinking,fridovich2018fastslow,graves2016adaptive}.
Large portions of observations are familiar, and the action trajectories given such observations are predictable. Under such circumstances, a learned, reactive policy with low latency can already provide reliable actions; repeatedly invoking a full planner for these routine observations offers little benefit relative to its cost.
Planning becomes more valuable when the agent encounters unfamiliar observations, disturbances, or situations where its learned behavior is less reliable.
However, many test-time planners that utilize a learned world model devote the same computational budget to every timestep, ignoring the fact that not all decisions require planning.

A similar distinction arises in human decision-making.
Dual-process theory \citep{kahneman2011thinking,daw2005uncertainty,botvinick2009planning} describes two complementary modes of thought: (i) a fast, intuitive \emph{System 1} that handles familiar situations through learned experience, and (ii) a slower, deliberative \emph{System 2} that reasons through complex or novel situations.
Humans naturally alternate between both systems for effective decision-making. 

\begin{figure}[t]
    \centering
    \includegraphics[width=0.9\linewidth]{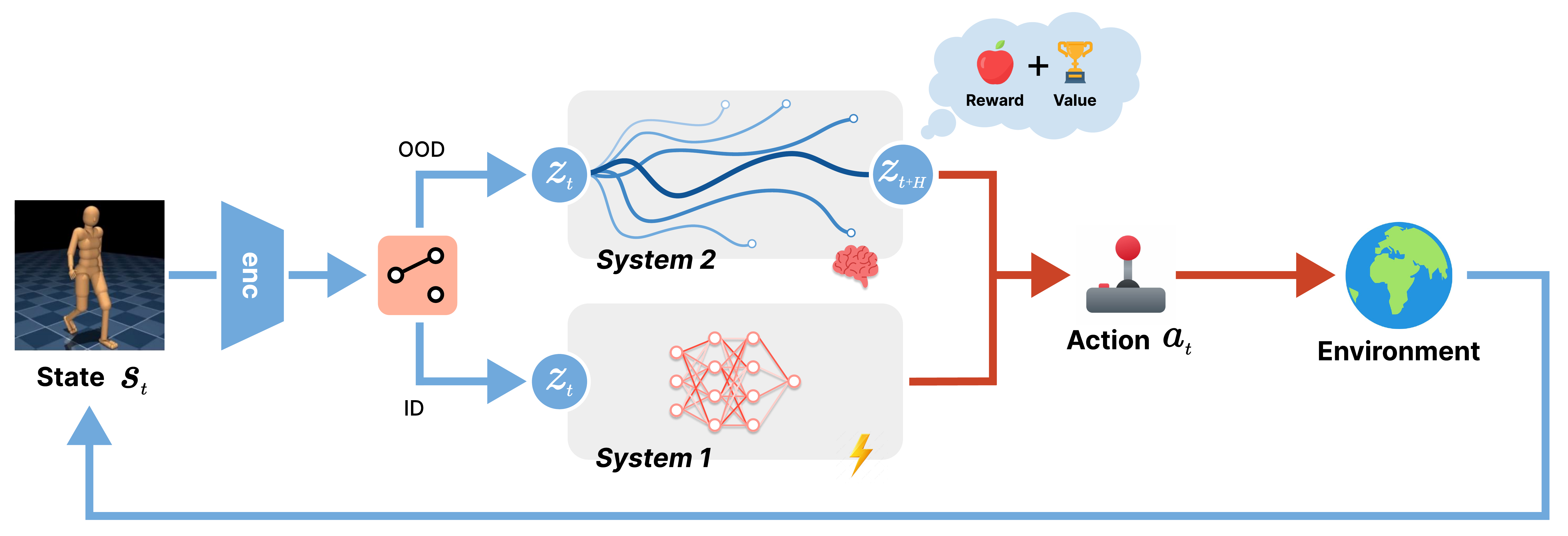}
    \caption{\textbf{Overview of Fast-TD-MPC.} Fast-TD-MPC uses TD-MPC2's \gls{mppi} planner as System~2 (the slow, robust fallback), a compact amortized MLP trained via planner imitation as System~1 (the fast, reactive policy), and a lightweight OOD detector as the gating mechanism. At each decision step, the OOD detector evaluates the current state: in-distribution (ID) states are routed to the fast and compact System~1, while out-of-distribution (OOD) states fall back to System~2 (TD-MPC2's \gls{mppi} planner). Since the detector itself costs negligible computation, the effective per-step latency scales linearly with the fraction of timesteps that require planning.}
    \label{fig:Fast-TD-MPC_algo_figure}
    \vspace{-4pt}
\end{figure}

Motivated by this view, we propose \textbf{Fast-TD-MPC}, a framework that dynamically switches between fast reactive execution and test-time planning.
Applied to TD-MPC2 \citep{hansen2024tdmpc2}, Fast-TD-MPC uses the original \gls{mppi} planner as System 2 (the slow but robust fallback), a compact amortized neural network trained via planner imitation as System 1 (the fast, reflexive policy), and a lightweight out-of-distribution (OOD) detector as the gating mechanism (Figure~\ref{fig:Fast-TD-MPC_algo_figure}).
When the current observation is familiar, System 1 provides a reliable action at low latency; when the observation is OOD, System 2 is triggered to deliberate and plan.
Since the detector itself costs negligible computation, the effective per-step latency scales linearly with the fraction of timesteps that require planning.

Our contributions are threefold:
\begin{itemize}[leftmargin=*,itemsep=2pt,topsep=-2pt]
    \item We introduce \textbf{Fast-TD-MPC}, a framework for adaptive test-time planning in data-driven \gls{mpc}. Applied to TD-MPC2 \citep{hansen2024tdmpc2}, it combines the original \gls{mppi} planner as a robust System~2 with a compact policy trained through planner imitation as a fast System~1, gated by a lightweight OOD detector, without modifying the underlying world model or planner.

    \item We conduct extensive evaluations across \textbf{103 continuous-control tasks} spanning four domains: DMControl \citep{tassa2018dmcontrol} (physics simulated locomotion and manipulation), Meta-World \citep{yu2019metaworld} (multi-object manipulation), ManiSkill2 \citep{gu2023maniskill2} (robotic manipulation), and MyoSuite \citep{caggiano2022myosuite} (physiologically realistic musculoskeletal motor control).
    Fast-TD-MPC achieves up to ${\sim}4\times$ inference speedup (${\sim}2.7\times$ on average) with ${<}10\%$ performance drop on 3 of 4 domains, using a single set of hyperparameters across all tasks (Figure~\ref{fig:main_pareto_clean}).

    \item We show that faster decision-making does not come at the expense of robustness. Under five disturbance types spanning three categories, Fast-TD-MPC selectively falls back to planning when the agent encounters out-of-distribution states, retaining 95\% of TD-MPC2's perturbed return on DMControl at half its latency.

\end{itemize}

%% file: sections/method.tex
\section{Preliminaries}
\label{sec:preliminaries}

\paragraph{Problem formulation.}
We consider the standard infinite-horizon \gls{mdp} \citep{puterman1990mdp} setting, characterized by a tuple $(\mathcal{S}, \mathcal{A}, T, R, \gamma, p_0)$, where $s \in \mathcal{S}$ and $a \in \mathcal{A}$ are continuous states and actions, $T: \mathcal{S} \times \mathcal{A} \to \mathcal{S}$ is the transition (dynamics) function, $R: \mathcal{S} \times \mathcal{A} \to \mathbb{R}$ is a reward function associated with a particular task, $\gamma \in [0,1)$ is a discount factor, and $p_0$ is the initial state distribution. The goal is to derive a control policy $\pi: \mathcal{S} \to \mathcal{A}$ such that the expected discounted sum of rewards (return)
$ \mathop{\mathbb{E}}_{\pi}\left[\sum_{t=0}^{\infty} \gamma^t r_t\right], \quad r_t = \mathcal{R}(s_t, \pi(s_t)) $
is maximized. 
In practice, each step must complete within the environment's control cycle; slow per-step computation incurs stale-action errors and degraded return.

\paragraph{TD-MPC2.}
Our work builds on TD-MPC2 \citep{hansen2024tdmpc2}, an algorithm that combines test-time planning in a latent space using \gls{mpc} \citep{negenborn2005mpc} with actor-critic \gls{td} learning. TD-MPC2 consists of five components as follows ($\theta$ denotes all learnable parameters, $d_z$ the latent state dimensionality, $d_a$ the action dimensionality):
\begin{itemize}[leftmargin=*,noitemsep]
    \item An encoder $\mathcal{E}_{\theta}: \mathcal{S} \to \mathbb{R}^{d_z}$ mapping a state $s$ to its latent representation $z = \mathcal{E}_{\theta}(s)$.
    \item A latent dynamics model $\mathcal{D}_{\theta}: \mathbb{R}^{d_z} \times \mathcal{A} \to \mathbb{R}^{d_z}$, predicting the next latent $\hat{z}' = \mathcal{D}_{\theta}(z, a)$.
    \item A reward predictor $\mathcal{R}_{\theta}: \mathbb{R}^{d_z} \times \mathcal{A} \to \mathbb{R}$, computing expected rewards $\hat{r} = \mathcal{R}_{\theta}(z, a)$.
    % via discrete regression in a log-transformed space.
    \item A policy prior $\pi_{\theta}: \mathbb{R}^{d_z} \to \mathcal{A}$, a stochastic maximum-entropy policy that guides planning, from which actions $a \sim \pi_{\theta}(\cdot \mid z)$ are sampled.
    \item A critic $Q_{\theta}: \mathbb{R}^{d_z} \times \mathcal{A} \to \mathbb{R}$, estimating state-action values $Q_{\theta}(z, a) \approx Q^{\pi_{\theta}}(s, a)$. TD-MPC2 trains an ensemble of $N_Q = 5$ critics and computes TD-targets as the minimum of two randomly subsampled target critics to mitigate value overestimation.
\end{itemize}
The world model components $\mathcal{E}_{\theta}$, $\mathcal{D}_{\theta}$, $\mathcal{R}_{\theta}$, and $Q_{\theta}$ are jointly optimized by minimizing a combined objective over sub-trajectories of horizon $H$ sampled from a replay buffer $\mathcal{B}$:
\begin{equation}
    \mathcal{L}(\theta) \doteq \mathop{\mathbb{E}}_{(s, a, r, s')_{0:H} \sim \mathcal{B}} \left[\sum_{t=0}^{H} \lambda^t \left( \underbrace{\| \hat{z}'_t - \mathrm{sg}(\mathcal{E}_{\theta^{-}}(s'_t)) \|^2}_{\text{joint-embedding prediction}} + \underbrace{\mathrm{CE}(\hat{r}_t, r_t)}_{\text{reward prediction}} + \underbrace{\mathrm{CE}(\hat{q}_t, q_t)}_{\text{value prediction}} \right)\right],
\end{equation}
where $\hat{z}_t$ are recurrently predicted latents ($\hat{z}_0 = \mathcal{E}_{\theta}(s_0)$, $\hat{z}'_t = \mathcal{D}_{\theta}(\hat{z}_t, a_t)$), $\mathrm{sg}$ is the stop-gradient operator, $\theta^{-}$ are target parameters updated as an exponential moving average (EMA) of $\theta$, $\lambda \in (0, 1]$ is a temporal coefficient that down-weights distant predictions, $q_t \doteq r_t + \gamma \bar{Q}(z'_t, \pi_{\theta}(z'_t))$ is the TD-target with $\bar{Q}$ being an EMA of the critic ensemble, and $\mathrm{CE}$ denotes cross-entropy with soft two-hot targets in a symlog-transformed discrete space.

The policy prior $\pi_{\theta}$ is trained independently by maximizing:
\begin{equation}
    \mathcal{L}_{\pi}(\theta) \doteq \mathop{\mathbb{E}}_{(s, a)_{0:H} \sim \mathcal{B}} \left[\sum_{t=0}^{H} \lambda^t \left[\alpha \, Q_{\theta}(z_t, \pi_{\theta}(z_t)) + \beta \, \mathcal{H}\!\left(\pi_{\theta}(\cdot \mid z_t)\right)\right]\right],
\end{equation}
where $\mathcal{H}$ is the policy entropy and $\alpha, \beta$ are coefficients balanced via moving statistics to prevent premature entropy collapse. Gradients of $\mathcal{L}_{\pi}$ are taken only with respect to the parameters of $\pi_{\theta}$.

At inference time, TD-MPC2 plans in the latent space using \gls{mppi} \citep{williams2015mppi}. Let $\mu, \sigma \in \mathbb{R}^{H \times d_a}$ be the parameters of a time-dependent multivariate Gaussian with diagonal covariance. Given the current state $s_t$ of a decision step $t$, the encoder produces $z_t = \mathcal{E}_{\theta}(s_t)$, and \gls{mppi} iteratively refines the parameters $(\mu, \sigma)$ over action sequences of horizon $H$ by (i) sampling $N$ candidate trajectories, including a fraction from the policy prior $\pi_{\theta}$; (ii) evaluating each via latent rollout and terminal value bootstrapping:
\begin{equation}
    \hat{J} = \gamma^{H} Q_{\theta}(z_H, \pi_{\theta}(z_H)) + \sum_{h=0}^{H-1} \gamma^{h} \mathcal{R}_{\theta}(z_h, a_h);
\end{equation}
and (iii) updating $(\mu, \sigma)$ via an importance weighted average over top-$k$ elite trajectories, yielding refined parameters $(\mu^{*}_t, \sigma^{*}_t)$. Only the first action $a_t \sim \mathcal{N}(\mu^{*}_t, \sigma^{*2}_t)$ is executed, and the process repeats the next decision step, with $\mu^{*}_t$ shifted by one position to warm-start the next iteration's sampling. We refer readers to their paper for exact algorithmic details.

\section{Fast-TD-MPC: A Framework for Adaptive Test-Time Planning }
\label{sec:method}

We study accelerating TD-MPC2 inference while preserving its robustness to states outside the expert distribution. Given a pre-trained TD-MPC2 world model (with frozen parameters $\theta$), our goal is to route each decision step to the fastest system capable of handling it, \emph{without} significantly degrading return. Fast-TD-MPC consists of three components, all frozen prior to deployment (Figure~\ref{fig:Fast-TD-MPC_algo_figure}): (i) an amortized policy (System 1), (ii) the \gls{mppi} planner (System 2), and (iii) an OOD gate.

\paragraph{System 1: Amortized Policy.}
\emph{Planner amortization} reduces the per-step cost of \gls{mppi} by training a lightweight feed-forward policy $\pi_{\text{S1}}$ to imitate the planner. Given expert demonstrations $\mathcal{D}_{\text{exp}} = \{(z_t, a^{*}_t)\}$ collected via planner rollouts, we minimize the imitation objective:
$\min_\phi \; \mathop{\mathbb{E}}_{(z, a^{*}) \sim \mathcal{D}_{\text{exp}}} \left[ \ell\!\left(\pi_{\text{S1}}(z;\phi),\, a^{*}\right) \right]$,
where $\phi$ are the parameters of $\pi_{\text{S1}}$ and $\ell$ is an L1 loss, with $|\hat{a} - a^{*}|$ denoting the element-wise absolute difference summed over $d_a$. See \appdxref{apdx:implementation_details} for training and architectural details. While this reduces inference to a single forward pass, the amortized policy cannot recover from states outside the expert distribution. We address this via OOD detection.

\paragraph{System 2: MPPI Planner.}
We use TD-MPC2's world model and \gls{mppi} planning algorithm (Section~\ref{sec:preliminaries}) as-is. When the gate switches from System~1 to System~2 after $K$ timesteps, the warm-start buffer is $K$-step stale. We explicitly zero the buffer on System~1 to System~2 transitions, forcing a cold start from the policy prior $\pi_\theta$, which provides fresh forward-looking candidates via the world model. The planner resumes normal warm-starting on subsequent consecutive System 2 steps.

\paragraph{OOD Gating.}
To address the failure mode of the amortized policy, we employ out-of-distribution (OOD) detection \citep{hendrycks2017baseline,lee2018mahalanobis} in the latent space. We employ Mahalanobis distance \citep{mahalanobis1936distance} as the scoring function:
$g(z_t) = (z_t - \bar{z})^{\top} \Sigma_{\text{reg}}^{-1} (z_t - \bar{z})$,
where $\bar{z}$ and $\Sigma$ are the mean and full covariance of in-distribution (ID) latents, fitted offline with $\ell_2$ shrinkage ($\Sigma_{\text{reg}} = \Sigma + \lambda \mathbf{I}$, $\lambda = 0.01$). To construct the ID set, we roll out $\pi_{\text{S1}}$ and collect all visited latents $\{z_t\}$. We consider two labeling modes: \emph{(i) theoretical}, where all visited states are treated as ID, and \emph{(ii) reward-gated}, where a state is labeled ID only if its per-step rewards meet or exceed the average per-step counterparts of TD-MPC2 expert rollouts at the same timestep. The gating is then:
\begin{equation}
\label{eq:gating}
a_t = \begin{cases} \pi_{\text{S1}}(z_t; \phi) & \text{if } g(z_t) \leq \tau, \\ \pi_{\text{S2}}(z_t; \theta) & \text{if } g(z_t) > \tau, \end{cases}
\end{equation}
where $\tau$ controls the speed--robustness trade-off. For \emph{theoretical} mode, $\tau$ defaults to the median\footnote{Under Gaussian assumptions, the squared Mahalanobis distance follows a $\chi^2$ distribution \citep{mahalanobis1936distance}, which is right-skewed; the median is therefore more representative of the typical score than the mean.} of all scores on a held-out validation set. For \emph{reward-gated} mode, $\tau$ defaults to the midpoint between the median scores of the ID and OOD partitions. We additionally sweep multiple percentile values and report the full Pareto frontier in Section~\ref{sec:rq1}.

\paragraph{Latency Profile.}
Since $g(z_t)$ is a single matrix-vector product, the overhead of OOD detection is negligible. The effective per-step latency is: 
$t_{\text{total}} \approx \mathcal{E}_\theta + g + \rho \cdot \pi_{\text{S2}} + (1{-}\rho) \cdot \pi_{\text{S1}} + t_{\text{env}}$, 
where $\rho$ is the S2 invocation rate, $\pi_{\text{S2}} \gg \pi_{\text{S1}}$, and $\mathcal{E}_\theta$, $g$, and $t_{\text{env}}$ are incurred at every step. As $\rho \to 0$, the system approaches the speed of a purely amortized policy; as $\rho \to 1$, it approaches the robustness of full planning. The threshold $\tau$ directly controls this trade-off.

\paragraph{Algorithm.}
Algorithm~\ref{alg:fast_tdmpc} summarizes the full inference procedure. 
All components are computed offline and frozen at deployment.

%% file: sections/experiments.tex
\section{Experiments}
\label{sec:experiments}

\begin{figure}[t]
    \centering
    \begin{minipage}[t]{0.22\linewidth}
        \centering
        \includegraphics[width=\linewidth]{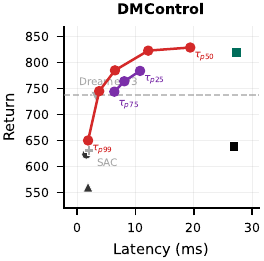}
    \end{minipage}\hfill
    \begin{minipage}[t]{0.22\linewidth}
        \centering
        \includegraphics[width=\linewidth]{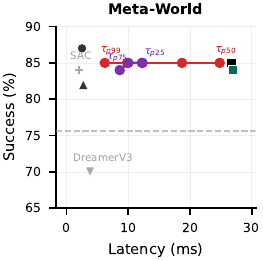}
    \end{minipage}\hfill
    \begin{minipage}[t]{0.22\linewidth}
        \centering
        \includegraphics[width=\linewidth]{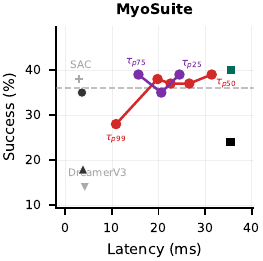}
    \end{minipage}\hfill
    \begin{minipage}[t]{0.22\linewidth}
        \centering
        \includegraphics[width=\linewidth]{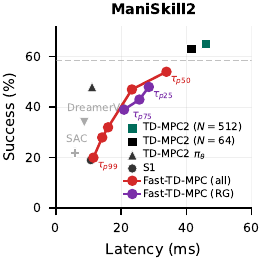}
    \end{minipage}
    \caption{\textbf{Pareto frontier.} Return/success rate vs.\ per-step latency across 4 domains. Fast-TD-MPC (colored) traces the Pareto frontier across varying OOD thresholds, matching or surpassing baselines at comparable latency on DMControl, Meta-World, and ManiSkill2. Dashed line marks 90\% of TD-MPC2's performance. DreamerV3 and SAC performance referenced from \citet{hansen2024tdmpc2}.}
    \label{fig:main_pareto_clean}
    \vspace{-3pt}
\end{figure}

\begin{table}[t]
\scriptsize
\setlength{\tabcolsep}{4.5pt}
\newcommand{\pmci}[2]{$#1_{\scriptscriptstyle \pm #2}$}
\caption{
\textbf{Benchmark results across 4 domains (103 tasks, 3 seeds each).} Mean and 95\% CI across all tasks in each domain. \textsuperscript{$\dag$}Return and success rate for DreamerV3 and SAC are referenced from \citet{hansen2024tdmpc2}. \emph{all} denotes the \emph{theoretical} OOD mode; \emph{RG} denotes the \emph{reward-gated} mode. See \appdxref{apdx:full_results} for full results including S2 invocation rate and peak GPU memory usage. %TODO
}
\label{tab:main_results}
\begin{center}
\begin{tabular}{l *{8}{r}}
\toprule
& \multicolumn{2}{c}{\textbf{DMControl}} & \multicolumn{2}{c}{\textbf{Meta-World}} & \multicolumn{2}{c}{\textbf{MyoSuite}} & \multicolumn{2}{c}{\textbf{ManiSkill2}} \\
\cmidrule(lr){2-3} \cmidrule(lr){4-5} \cmidrule(lr){6-7} \cmidrule(lr){8-9}
\textbf{Method} & \textbf{Return} & \textbf{Lat.} & \textbf{Succ.} & \textbf{Lat.} & \textbf{Succ.} & \textbf{Lat.} & \textbf{Succ.} & \textbf{Lat.} \\
 &  & {(ms)} & {(\%)} & {(ms)} & {(\%)} & {(ms)} & {(\%)} & {(ms)} \\
\midrule
DreamerV3\textsuperscript{$\dag$} & \pmci{735}{93} & \pmci{3.2}{.1} & \pmci{70}{8} & \pmci{3.8}{.1} & \pmci{14}{17} & \pmci{4.2}{.1} & \pmci{34}{56} & \pmci{8.7}{.7} \\
SAC\textsuperscript{$\dag$} & \pmci{631}{108} & \pmci{2.0}{.4} & \pmci{84}{7} & \pmci{2.0}{.2} & \pmci{38}{20} & \pmci{2.9}{.5} & \pmci{22}{54} & \pmci{5.8}{.7} \\
\midrule
S2-alone (TD-MPC2) & \pmci{820}{55} & \pmci{27.3}{1.1} & \pmci{84}{9} & \pmci{26.9}{.2} & \pmci{40}{20} & \pmci{35.5}{.4} & \pmci{65}{54} & \pmci{45.9}{4.8} \\
S2-alone ($N{=}64$) & \pmci{639}{100} & \pmci{26.9}{1.1} & \pmci{85}{7} & \pmci{26.7}{.2} & \pmci{24}{16} & \pmci{35.4}{.4} & \pmci{63}{53} & \pmci{41.5}{3.4} \\
S2 policy-only & \pmci{560}{105} & \pmci{1.9}{.3} & \pmci{82}{8} & \pmci{2.7}{.2} & \pmci{18}{15} & \pmci{3.8}{.4} & \pmci{48}{56} & \pmci{11.1}{2.5} \\
S1 alone & \pmci{623}{91} & \pmci{1.6}{.2} & \pmci{87}{8} & \pmci{2.5}{.2} & \pmci{35}{16} & \pmci{3.6}{.3} & \pmci{19}{39} & \pmci{10.7}{3.9} \\
\midrule
Round-robin (S2 50\%) & \pmci{813}{57} & \pmci{14.4}{.7} & \pmci{86}{7} & \pmci{14.7}{.2} & \pmci{45}{21} & \pmci{19.3}{.3} & \pmci{60}{44} & \pmci{28.4}{5.6} \\
Round-robin (S2 33\%) & \pmci{782}{66} & \pmci{10.1}{.5} & \pmci{86}{8} & \pmci{10.7}{.2} & \pmci{41}{18} & \pmci{14.1}{.3} & \pmci{42}{50} & \pmci{22.3}{3.8} \\
Round-robin (S2 20\%) & \pmci{743}{74} & \pmci{6.6}{.4} & \pmci{85}{8} & \pmci{7.3}{.2} & \pmci{40}{22} & \pmci{9.8}{.3} & \pmci{38}{48} & \pmci{17.6}{3.9} \\
Round-robin (S2 10\%) & \pmci{699}{83} & \pmci{4.0}{.3} & \pmci{86}{8} & \pmci{4.9}{.2} & \pmci{40}{21} & \pmci{6.6}{.3} & \pmci{28}{46} & \pmci{14.5}{4.7} \\
\midrule
Fast-TD-MPC (\emph{all}, $\tau_{p50}$) & \pmci{829}{55} & \pmci{19.4}{2.4} & \pmci{85}{8} & \pmci{24.8}{1.1} & \pmci{39}{21} & \pmci{31.4}{3.6} & \pmci{54}{55} & \pmci{33.9}{12.1} \\
Fast-TD-MPC (\emph{all}, $\tau_{p75}$) & \pmci{823}{55} & \pmci{12.2}{2.7} & \pmci{85}{8} & \pmci{18.7}{1.5} & \pmci{37}{21} & \pmci{26.6}{5.8} & \pmci{47}{54} & \pmci{23.3}{10.7} \\
Fast-TD-MPC (\emph{all}, $\tau_{p90}$) & \pmci{785}{58} & \pmci{6.5}{2.0} & \pmci{85}{8} & \pmci{12.2}{1.6} & \pmci{37}{21} & \pmci{22.6}{7.6} & \pmci{32}{42} & \pmci{16.0}{5.2} \\
Fast-TD-MPC (\emph{all}, $\tau_{p95}$) & \pmci{745}{74} & \pmci{3.8}{1.4} & \pmci{85}{8} & \pmci{9.8}{1.5} & \pmci{38}{23} & \pmci{19.8}{7.6} & \pmci{28}{46} & \pmci{14.2}{3.9} \\
Fast-TD-MPC (\emph{all}, $\tau_{p99}$) & \pmci{650}{92} & \pmci{1.9}{.3} & \pmci{85}{8} & \pmci{6.2}{1.3} & \pmci{28}{17} & \pmci{10.9}{5.8} & \pmci{20}{39} & \pmci{11.5}{3.9} \\
\midrule
Fast-TD-MPC (\emph{RG}, $\tau_{p25}$) & \pmci{784}{68} & \pmci{10.8}{2.3} & \pmci{85}{8} & \pmci{12.3}{1.7} & \pmci{39}{22} & \pmci{24.5}{4.6} & \pmci{48}{43} & \pmci{28.5}{20.9} \\
Fast-TD-MPC (\emph{RG}, $\tau_{p50}$) & \pmci{764}{70} & \pmci{8.1}{1.8} & \pmci{85}{8} & \pmci{10.0}{1.5} & \pmci{35}{21} & \pmci{20.6}{5.2} & \pmci{43}{43} & \pmci{25.6}{18.9} \\
Fast-TD-MPC (\emph{RG}, $\tau_{p75}$) & \pmci{744}{74} & \pmci{6.4}{1.5} & \pmci{84}{8} & \pmci{8.6}{1.5} & \pmci{39}{21} & \pmci{15.7}{4.7} & \pmci{39}{51} & \pmci{20.9}{10.8} \\
\bottomrule
\end{tabular}
\end{center}
\vspace{-1.0em}
\end{table}

% ─── RQs ───
We seek to address the following research questions:
\begin{itemize}[leftmargin=*,itemsep=3pt]
    \item \textit{\textbf{RQ1:} Can Fast-TD-MPC achieve substantial inference speedup over TD-MPC2 without significant degradation in return or success rate across diverse continuous control tasks?}
    \item \textit{\textbf{RQ2:} Can Fast-TD-MPC preserve the robustness of the original TD-MPC2 system under various disturbances?}
\end{itemize}
We additionally analyze the speed-robustness trade-off via $\tau$ and ablate the \gls{mppi} planner warm-start strategy and OOD distance metric.

% ─── 4.1 Setup ───
\subsection{Experimental Setup}

We evaluate Fast-TD-MPC across 103 continuous control tasks spanning 4 domains: DMControl \citep{tassa2018dmcontrol}, Meta-World \citep{yu2019metaworld}, ManiSkill2 \citep{gu2023maniskill2}, and MyoSuite \citep{caggiano2022myosuite}, with action spaces up to $\mathcal{A} \in \mathbb{R}^{39}$ and diverse tasks from manipulation to complex locomotion. We consider the single-task setting with 3 seeds per task. 
All Fast-TD-MPC components are fitted independently for each task-seed pair using a single set of hyperparameters across all tasks; no per-task tuning done. 
See \appdxref{apdx:task_domains} and \textbf{\ref{apdx:implementation_details}} for task and hyperparameter details.

\paragraph{Baselines.}
We compare against 5 TD-MPC2-derived baselines:
(i) \textbf{S2-alone} (full TD-MPC2, $N{=}512$),
(ii) \textbf{S2-reduced} ($N{=}64$),
(iii) \textbf{S2 policy-only} (no MPPI),
(iv) \textbf{S1 alone} (amortized MLP),
(v) \textbf{Round-robin} (S2 invoked at a fixed rate of 50\%, 33\%, 20\%, or 10\%).
We evaluate 2 OOD modes (\emph{theoretical} and \emph{reward-gated}) across multiple $\tau$ percentiles, and reference DreamerV3 \citep{hafner2023dreamerv3} and SAC \citep{haarnoja2018sac} numbers from \citet{hansen2024tdmpc2}.
We follow \citet{hansen2024tdmpc2}'s evaluation setup, and report return/success rate, per-step latency (ms), S2 invocation rate ($\rho$), as well as peak GPU memory. Refer \appdxref{apdx:evaluation_setup_and_hardware} for hardware and evaluation details.

% ─── 4.2 RQ1: Speedup vs Performance ───
% RQ: Can Fast-TD-MPC achieve significant speedup without significant performance drop?

\subsection{Speedup vs. Performance (\textit{\textbf{RQ1}})}
\label{sec:rq1}

We present aggregate results in Tab.~\ref{tab:main_results} and Fig.~\ref{fig:main_pareto_clean}, with per-component latency breakdowns in Tab.~\ref{tab:runtime_profile}.

\paragraph{Pareto-optimal speed--performance trade-off.}
At $\tau_{p50}$, Fast-TD-MPC (\emph{all}) matches TD-MPC2 within the 95\% CI across all domains (829 vs.\ 820 on DMControl, 85\% vs.\ 84\% on Meta-World) while reducing latency by ${\sim}1.3$--$1.4\times$ (27.3\,ms $\to$ 19.4\,ms on DMControl, 45.9\,ms $\to$ 33.9\,ms on ManiSkill2). Varying $\tau$ traces a Pareto frontier: 
at $\tau_{p75}$, speedup reaches ${\sim}2\times$ (27.3\,ms $\to$ 12.2\,ms on DMControl, and 45.9\,ms $\to$ 23.3\,ms on ManiSkill2); 
at $\tau_{p90}$, Fast-TD-MPC achieves ${\sim}2.7\times$ average speedup (up to $4.2\times$ on DMControl) with ${<}10\%$ performance drop on 3 of 4 domains, retaining 96\% of TD-MPC2's return on DMControl, matching its success rate on Meta-World, and retaining 93\% on MyoSuite.
Beyond $p_{90}$, performance degrades sharply: at $\tau_{p99}$, DMControl return drops to 650 (${\sim}20\%$ loss) and ManiSkill2 success to 20\%, as the gate routes nearly all steps to System 1, reducing to the S1-alone regime.

\vspace{-4pt}

\paragraph{State-aware gating vs.\ fixed scheduling.}
Round-robin (RR) invokes System~2 at a fixed rate regardless of state. On DMControl, gating dominates every RR operating point at matched latency, e.g., 785 return at 6.5\,ms ($\tau_{p90}$) vs.\ 743 at 6.6\,ms (RR 20\%), and 745 at 3.8\,ms ($\tau_{p95}$) vs.\ 699 at 4.0\,ms (RR 10\%). On ManiSkill2, the two are comparable: gating is ahead at intermediate latency, with 47\% at 23.3\,ms ($\tau_{p75}$) vs.\ 42\% at 22.3\,ms (RR 33\%), while RR 50\% is stronger at the high-latency end (60\% at 28.4\,ms vs.\ 54\% at 33.9\,ms for $\tau_{p50}$). On Meta-World, all scheduling variants lie within 2 points (84--86\%), indicating saturation. On MyoSuite, RR (S2 10\%) recovers TD-MPC2's success (40\%) at 6.6\,ms, whereas the gate still invokes S2 on 60\% of steps at $\tau_{p90}$. Thus, \emph{which} timesteps are planned matters most when only a small subset of states requires planning; otherwise, fixed scheduling is a simpler alternative.

\vspace{-4pt}

\paragraph{OOD mode comparison.}
The \emph{reward-gated} mode matches the \emph{theoretical} mode's performance at lower latency across multiple domains. On Meta-World, \emph{RG} $\tau_{p50}$ achieves 85\% at 10.0\,ms vs.\ \emph{all} $\tau_{p90}$'s 85\% at 12.2\,ms. On MyoSuite, \emph{RG} $\tau_{p75}$ achieves 39\% at 15.7\,ms, matching \emph{all} $\tau_{p50}$'s 39\% at 31.4\,ms and exceeding \emph{all} $\tau_{p90}$'s 37\% at 22.6\,ms. This suggests reward-gating's tighter ID distribution produces a more precise threshold, reducing unnecessary S2 invocations on states System~1 can reliably handle.

\vspace{-4pt}

\paragraph{Reference baselines.}
Both DreamerV3 and SAC are faster than TD-MPC2 but achieve lower returns, particularly on DMControl (735 and 631 vs.\ 820) and MyoSuite (14\% and 38\% vs.\ 40\%). Among TD-MPC2-derived baselines, reducing the planner's sample budget ($N{=}512$ $\to$ $N{=}64$) degrades return on DMControl by 22\% and MyoSuite by 40\%, while providing little latency reduction, as latency is dominated by fixed GPU overhead rather than sample count. S1 alone achieves the lowest latency but degrades significantly, particularly on ManiSkill2 (19\% vs.\ 65\%) and DMControl (623 vs.\ 820), confirming that pure amortization without a fallback is insufficient. The policy-only variant (no MPPI) is similarly fast but performs worse than S1 on most domains, confirming that supervised distillation from \gls{mppi} yields a stronger reactive policy than the TD-MPC2 actor alone.

\vspace{-4pt}

\paragraph{Runtime breakdown.}
Table~\ref{tab:runtime_profile} details the per-component latency breakdown. 
The OOD gate adds negligible overhead ($\sim$0.2 ms), while System~1 is $\sim$65$\times$ cheaper per invocation than System~2 ($\sim$0.4 ms vs.\ $\sim$26 ms on H100).
At $\tau_{p50}$, $\rho$ adapts to domain difficulty (81\% on H100 vs.\ 66\% on A100), and peak GPU memory remains bounded by S2-alone. 

\vspace{-4pt}

\begin{figure}[t]
    \centering
    \begin{minipage}[t]{0.22\linewidth}
        \centering
        \includegraphics[width=\linewidth]{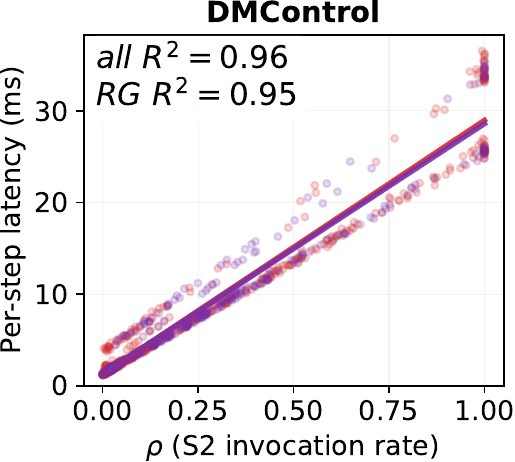}
    \end{minipage}\hfill
    \begin{minipage}[t]{0.22\linewidth}
        \centering
        \includegraphics[width=\linewidth]{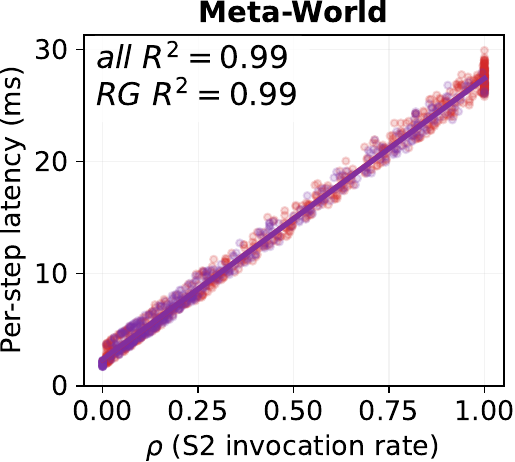}
    \end{minipage}\hfill
    \begin{minipage}[t]{0.22\linewidth}
        \centering
        \includegraphics[width=\linewidth]{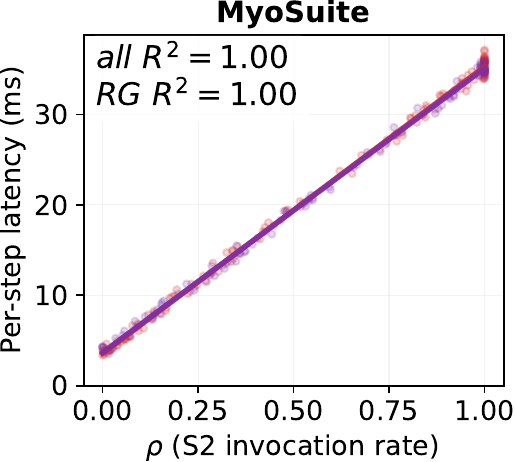}
    \end{minipage}\hfill
    \begin{minipage}[t]{0.22\linewidth}
        \centering
        \includegraphics[width=\linewidth]{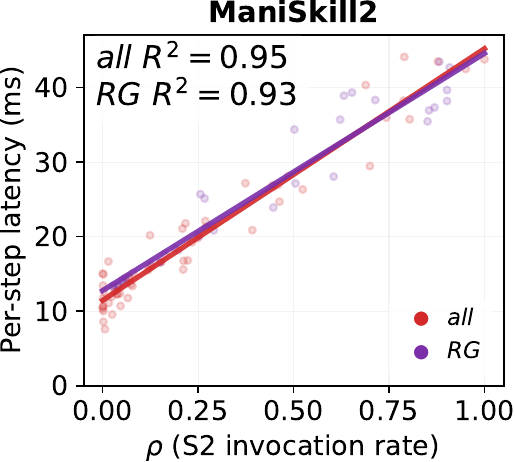}
    \end{minipage}
    \caption{\textbf{Per-step latency vs.\ S2 invocation rate ($\rho$).} Each point is a task-seed configuration; red denotes the \emph{theoretical} (\emph{all}) OOD mode, purple denotes the \emph{reward-gated} (\emph{RG}) mode. The linear relationship ($R^2 \geq 0.93$) confirms that per-step latency scales linearly with $\rho$.}
    \label{fig:latency_vs_rho}
    \vspace{-4pt}
\end{figure}

\paragraph{Latency scales linearly with planning frequency.}
Figure~\ref{fig:latency_vs_rho} confirms that per-step latency scales linearly with $\rho$ across all domains ($R^2 \geq 0.93$), validating the runtime model: since the OOD detector costs negligible computation (${\sim}0.2$\,ms), total latency is determined by the fraction of timesteps routed to System~2. Both \emph{all} and \emph{RG} modes fall on the same line, confirming that $\tau$ controls only the operating point along this line, not the per-step cost structure.

\vspace{-4pt}

\begin{table}[t]
\scriptsize
\caption{\textbf{Per-component latency breakdown by hardware group and peak GPU memory.} Results are averaged within hardware groups: DMControl, Meta-World, and MyoSuite on H100; ManiSkill2 on A100. $\pi_{\text{S1}}$ and $\pi_{\text{S2}}$ report per-invocation cost (zero entries excluded), $\rho$ is the S2 invocation rate, and $t_{\text{total}} \approx \mathcal{E}_\theta + g + \rho\pi_{\text{S2}} + (1{-}\rho)\pi_{\text{S1}} + t_{\text{env}}$.}
\label{tab:runtime_profile}
\setlength{\tabcolsep}{3.5pt}
\newcommand{\pmci}[2]{$#1_{\scriptscriptstyle \pm #2}$}

\begin{center}
\begin{tabular}{l rrrrr rr rr}
\toprule
& \multicolumn{4}{c}{\textbf{Policy components (ms)}} & & \textbf{Total} & $\rho$ & \textbf{GPU} \\
\cmidrule(lr){2-5}
\textbf{Method} & $\mathcal{E}_\theta$ & $g$ & $\pi_{\text{S1}}$ & $\pi_{\text{S2}}$ & \textbf{Env} & \textbf{(ms)} & (\%) & \textbf{(MB)} \\
\midrule
\multicolumn{9}{l}{\textit{H100} — DMControl + Meta-World + MyoSuite} \\
\midrule
S2-alone (TD-MPC2) & \pmci{0.31}{.00} & --- & --- & \pmci{26.2}{.6} & \pmci{1.4}{.2} & \pmci{27.9}{.7} & $100$ & \pmci{95}{0} \\
S1 alone & \pmci{0.38}{.00} & --- & \pmci{0.49}{.00} & --- & \pmci{1.4}{.2} & \pmci{2.3}{.2} & $0$ & \pmci{68}{0} \\
Fast-TD-MPC (\emph{all}, $\tau_{p50}$) & \pmci{0.35}{.00} & \pmci{0.17}{.00} & \pmci{0.40}{.01} & \pmci{26.2}{.5} & \pmci{1.4}{.2} & \pmci{23.3}{1.4} & \pmci{81}{4} & \pmci{90}{2} \\
Fast-TD-MPC (\emph{all}, $\tau_{p90}$) & \pmci{0.35}{.00} & \pmci{0.17}{.00} & \pmci{0.38}{.00} & \pmci{26.3}{.5} & \pmci{1.4}{.2} & \pmci{10.9}{1.6} & \pmci{33}{5} & \pmci{76}{2} \\
\midrule
\multicolumn{9}{l}{\textit{A100} — ManiSkill2} \\
\midrule
S2-alone (TD-MPC2) & \pmci{0.35}{.01} & --- & --- & \pmci{34.2}{.4} & \pmci{11.3}{4.4} & \pmci{45.8}{4.8} & $100$ & \pmci{70}{0} \\
S1 alone & \pmci{0.37}{.01} & --- & \pmci{0.65}{.01} & --- & \pmci{9.7}{3.9} & \pmci{10.7}{3.9} & $0$ & \pmci{42}{0} \\
Fast-TD-MPC (\emph{all}, $\tau_{p50}$) & \pmci{0.38}{.01} & \pmci{0.21}{.00} & \pmci{0.59}{.01} & \pmci{34.4}{.5} & \pmci{11.4}{4.8} & \pmci{34.9}{15.1} & \pmci{66}{37} & \pmci{63}{18} \\
Fast-TD-MPC (\emph{all}, $\tau_{p90}$) & \pmci{0.39}{.01} & \pmci{0.21}{.00} & \pmci{0.60}{.01} & \pmci{35.0}{.5} & \pmci{11.3}{2.6} & \pmci{17.0}{4.7} & \pmci{13}{9} & \pmci{46}{8} \\
\bottomrule
\end{tabular}
\end{center}
\vspace{-1.0em}
\end{table}

% ─── 4.3 RQ2: Robustness ───
% RQ: Can Fast-TD-MPC preserve TD-MPC2's robustness in noisy environments?

\subsection{Robustness Under Disturbances (\textit{\textbf{RQ2}})}
\label{sec:rq2}

\begin{figure}[t]
    \centering
    \begin{minipage}[t]{0.22\linewidth}
        \centering
        \includegraphics[width=\linewidth]{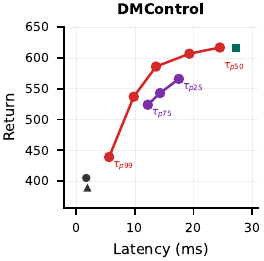}
    \end{minipage}\hfill
    \begin{minipage}[t]{0.22\linewidth}
        \centering
        \includegraphics[width=\linewidth]{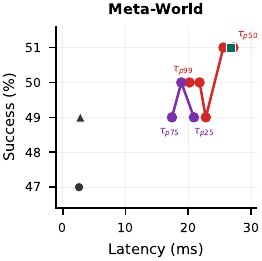}
    \end{minipage}\hfill
    \begin{minipage}[t]{0.22\linewidth}
        \centering
        \includegraphics[width=\linewidth]{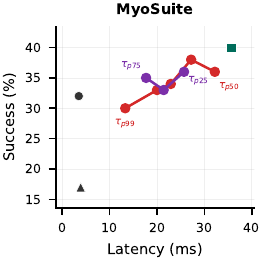}
    \end{minipage}\hfill
    \begin{minipage}[t]{0.22\linewidth}
        \centering
        \includegraphics[width=\linewidth]{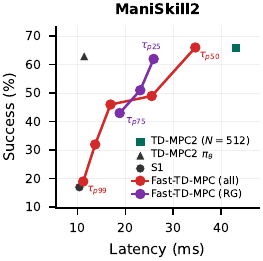}
    \end{minipage}
    \caption{\textbf{Pareto frontier under disturbances.} Return/success rate vs.\ per-step latency across 4 domains. Fast-TD-MPC (colored) matches or surpasses TD-MPC2 at lower latency on DMControl, Meta-World, and ManiSkill2, with a ${\sim}13\%$ performance drop on MyoSuite for up to ${\sim}2\times$ speedup.}
    \label{fig:pareto_mild}
    \vspace{-4pt}
\end{figure}

\begin{table}[t]
\scriptsize
\caption{
\textbf{Results of benchmarks with disturbances across 4 domains (102 tasks, 3 seeds each).} Mean and 95\% CI across all tasks in each domain. \emph{all} denotes \emph{theoretical} OOD mode; \emph{RG} denotes \emph{reward-gated} mode. 
See \appdxref{apdx:full_perturbation_results} for detailed results with all evaluation metrics. %TODO
}
\label{tab:perturbation_mild}
\setlength{\tabcolsep}{4.5pt}
\newcommand{\pmci}[2]{$#1_{\scriptscriptstyle \pm #2}$}
\begin{center}
\begin{tabular}{l *{8}{r}}
\toprule
& \multicolumn{2}{c}{\textbf{DMControl}} & \multicolumn{2}{c}{\textbf{Meta-World}} & \multicolumn{2}{c}{\textbf{MyoSuite}} & \multicolumn{2}{c}{\textbf{ManiSkill2}} \\
\cmidrule(lr){2-3} \cmidrule(lr){4-5} \cmidrule(lr){6-7} \cmidrule(lr){8-9}
\textbf{Method} & \textbf{Return} & \textbf{Lat.} & \textbf{Succ.} & \textbf{Lat.} & \textbf{Succ.} & \textbf{Lat.} & \textbf{Succ.} & \textbf{Lat.} \\
 &  & {(ms)} & {(\%)} & {(ms)} & {(\%)} & {(ms)} & {(\%)} & {(ms)} \\
\midrule
S2-alone (TD-MPC2) & \pmci{616}{104} & \pmci{27.2}{1.1} & \pmci{51}{10} & \pmci{26.8}{.2} & \pmci{40}{22} & \pmci{35.7}{.5} & \pmci{66}{75} & \pmci{43.1}{4.7} \\
S2 policy-only & \pmci{390}{102} & \pmci{1.9}{.3} & \pmci{49}{10} & \pmci{2.8}{.2} & \pmci{17}{15} & \pmci{3.9}{.4} & \pmci{63}{66} & \pmci{11.4}{5.7} \\
S1 alone & \pmci{405}{98} & \pmci{1.7}{.3} & \pmci{47}{10} & \pmci{2.6}{.2} & \pmci{32}{15} & \pmci{3.5}{.3} & \pmci{17}{52} & \pmci{10.4}{4.2} \\
\midrule
Round-robin (S2 50\%) & \pmci{577}{102} & \pmci{14.5}{.7} & \pmci{49}{11} & \pmci{14.7}{.2} & \pmci{37}{20} & \pmci{19.4}{.3} & \pmci{58}{89} & \pmci{27.1}{2.0} \\
\midrule
Fast-TD-MPC (\emph{all}, $\tau_{p50}$) & \pmci{617}{105} & \pmci{24.5}{1.9} & \pmci{51}{11} & \pmci{27.2}{.2} & \pmci{36}{20} & \pmci{32.2}{3.9} & \pmci{66}{80} & \pmci{34.6}{18.0} \\
Fast-TD-MPC (\emph{all}, $\tau_{p75}$) & \pmci{607}{103} & \pmci{19.3}{2.9} & \pmci{51}{11} & \pmci{25.6}{.9} & \pmci{38}{22} & \pmci{27.2}{5.9} & \pmci{49}{108} & \pmci{25.5}{22.3} \\
Fast-TD-MPC (\emph{all}, $\tau_{p90}$) & \pmci{586}{101} & \pmci{13.6}{3.0} & \pmci{49}{10} & \pmci{22.8}{1.6} & \pmci{34}{22} & \pmci{22.9}{7.8} & \pmci{46}{86} & \pmci{16.9}{10.5} \\
Fast-TD-MPC (\emph{all}, $\tau_{p95}$) & \pmci{537}{102} & \pmci{9.8}{2.6} & \pmci{50}{10} & \pmci{21.8}{1.8} & \pmci{33}{20} & \pmci{20.0}{7.8} & \pmci{32}{62} & \pmci{13.7}{5.2} \\
Fast-TD-MPC (\emph{all}, $\tau_{p99}$) & \pmci{439}{101} & \pmci{5.6}{2.0} & \pmci{50}{10} & \pmci{20.2}{2.0} & \pmci{30}{18} & \pmci{13.3}{7.8} & \pmci{19}{46} & \pmci{11.2}{3.1} \\
\midrule
Fast-TD-MPC (\emph{RG}, $\tau_{p25}$) & \pmci{566}{104} & \pmci{17.5}{2.6} & \pmci{49}{11} & \pmci{20.9}{1.9} & \pmci{36}{22} & \pmci{25.7}{4.7} & \pmci{62}{55} & \pmci{25.9}{25.6} \\
Fast-TD-MPC (\emph{RG}, $\tau_{p50}$) & \pmci{543}{102} & \pmci{14.3}{2.5} & \pmci{50}{10} & \pmci{18.9}{2.1} & \pmci{33}{22} & \pmci{21.4}{5.6} & \pmci{51}{61} & \pmci{23.1}{23.6} \\
Fast-TD-MPC (\emph{RG}, $\tau_{p75}$) & \pmci{524}{102} & \pmci{12.2}{2.3} & \pmci{49}{10} & \pmci{17.4}{2.2} & \pmci{35}{24} & \pmci{17.7}{5.3} & \pmci{43}{60} & \pmci{18.8}{14.7} \\
\bottomrule
\end{tabular}
\end{center}
\vspace{-1.5em}
\end{table}

We evaluate 102 of the 103 tasks under five types of externally applied disturbances, organized following the taxonomy of \citet{gu2025robustgym} into three groups:
(i) \emph{observation} (observation noise: additive Gaussian on $s_t$),
(ii) \emph{action} (action noise: additive Gaussian on $a_t$),
(iii) \emph{environment} (velocity kick: instantaneous impulse on $q_{\text{vel}}$; gravity shift: sustained rescaling of gravity; sustained force: continuous injection with probability $p{=}0.15$).
We report results for the \emph{combined} setting, which applies all five disturbances simultaneously. All disturbances are applied post-training at evaluation time with per-episode stochasticity. See \appdxref{apdx:env_details_perturb} for full details including the excluded task.

We compare against 
(i) \textbf{S2-alone}, 
(ii) \textbf{S2 policy-only}, 
(iii) \textbf{S1 alone}, and 
(iv) \textbf{Round-robin} (50\%), measuring absolute performance and degradation relative to the undisturbed setting. We present aggregate results across all 4 domains in Table~\ref{tab:perturbation_mild} and Figure~\ref{fig:pareto_mild}.

\vspace{-4pt}

\paragraph{Speed–performance trade-off holds under disturbances.}
At $\tau_{p90}$, Fast-TD-MPC (\emph{all}) retains 95\% of TD-MPC2's perturbed DMControl return (586 vs.\ 616) at $2\times$ speedup, while on ManiSkill2, $\tau_{p50}$ achieves 66\% success at 34.6\,ms vs.\ TD-MPC2's 66\% at 43.1\,ms. 
On Meta-World, all methods degrade to a similar level (47--51\%), leaving little room to
separate them.

\vspace{-4pt}

\begin{figure}[t]
    \centering
    \includegraphics[width=\linewidth]{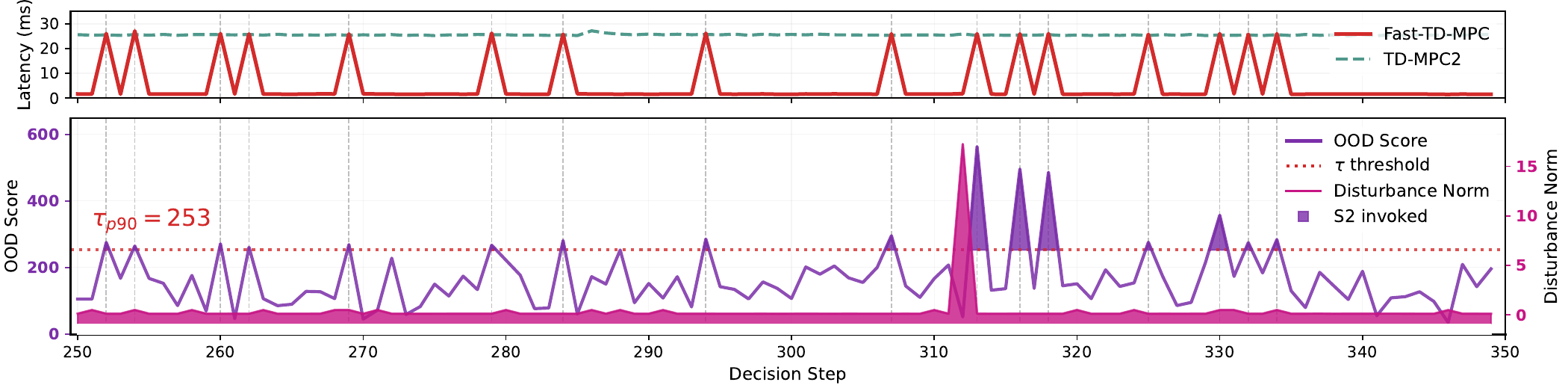}
    \vspace{-1em}
    \caption{\textbf{OOD gating dynamics.} Per-step OOD score and cumulative latency for a representative episode of \texttt{walker-walk-backwards} under combined disturbances (DMControl, $\tau_{p90}$). \textbf{Top:} Cumulative latency of Fast-TD-MPC (red) vs.\ TD-MPC2 (teal, dashed). \textbf{Bottom:} OOD score (purple, left axis) with $\tau_{p90}$ threshold (red dotted). Shaded regions and grey dashed lines mark timesteps where $g(z_t) > \tau$, indicating S2 invocation. Disturbance norm (pink, right axis) shows the total perturbation magnitude per step, with a spike at step 312 from the velocity kick.}
    \label{fig:ood_timeline}
    \vspace{-1em}
\end{figure}

\paragraph{OOD gating preserves robustness.}
Perturbations degrade TD-MPC2 substantially ($-25\%$ return on DMControl, $-39\%$ success on Meta-World), and S1 alone even more ($-35\%$ and $-46\%$), indicating that the amortized policy is more sensitive to distribution shift than the planner. At $\tau_{p50}$, where the gate routes most perturbed steps to the planner (88\% on DMControl, 99\% on Meta-World), Fast-TD-MPC (\emph{all}) tracks TD-MPC2 closely: 617 vs.\ 616 on DMControl, 51\% vs.\ 51\% on Meta-World, 66\% vs.\ 66\% on ManiSkill2, and 36\% vs.\ 40\% on MyoSuite. This robustness stems from adaptive routing rather than a fixed planning budget: at $\tau_{p90}$, the S2 invocation rate rises from 18\% to 45\% on DMControl and from 39\% to 82\% on Meta-World under perturbation (\appdxref{apdx:full_perturbation_results}). Figure~\ref{fig:ood_timeline} illustrates this at the step level: a velocity kick at step 312 pushes $g(z_t)$ above $\tau$, triggering S2 for several steps before the system resumes S1.

\vspace{-4pt}

\paragraph{State-aware gating vs.\ fixed scheduling under disturbances.}
Under perturbation, Fast-TD-MPC matches or slightly exceeds RR 50\% at a lower latency on DMControl (586 at 13.6\,ms vs.\ 577 at 14.5\,ms, $\tau_{p90}$) and ManiSkill2 (62\% at 25.9\,ms vs.\ 58\% at 27.1\,ms, \emph{RG} $\tau_{p25}$), although differences lie within the 95\% CIs. The key difference is adaptivity: the same $\tau_{p90}$ configuration runs at 6.5\,ms per step on clean DMControl episodes and increases planning only when perturbed, whereas RR 50\% spends 14.4\,ms per step in both conditions and degrades more ($813 \to 577$, $-29\%$, vs.\ $785 \to 586$, $-25\%$). On Meta-World and MyoSuite, RR is on par with or better than gating at matched latency.

\vspace{-4pt}

\paragraph{Reward-gated mode under disturbance.}
The reward-gated mode's efficiency advantage on Meta-World and MyoSuite (Section~\ref{sec:rq1}) persists under perturbation. On Meta-World, \emph{RG} $\tau_{p50}$ and $\tau_{p75}$ reach 49-50\% at 17.4-18.9\,ms, faster than any \emph{all} configuration (the fastest, $\tau_{p99}$, reaches 50\% at 20.2\,ms). On MyoSuite, \emph{RG} $\tau_{p75}$ achieves 35\% at 17.7\,ms, comparable to \emph{all} $\tau_{p50}$ (36\% at 32.2\,ms) and above \emph{all} $\tau_{p90}$ (34\% at 22.9\,ms). In both cases, gains come from fewer S2 invocations (e.g., 44\% vs.\ 60--89\% on MyoSuite), suggesting that reward-gating's tighter ID distribution produces a more precise threshold, reducing unnecessary S2 invocations on states System~1 can reliably handle.

% ─── 4.4 Ablations ───
\subsection{Ablations}
\label{sec:ablations}

\paragraph{MPPI Planner Warm-Start Strategy.}
We compare cold starting the \gls{mppi} planner against $K$-shifted warm start, which salvages the first $H{-}K$ steps of the stale buffer when $K < H{-}1$ and cold-starts otherwise. A paired t-test across all task-config comparisons on all 4 benchmark domains finds no significant difference ($p > 0.05$) in return ($t{=}1.31$, $p{=}0.19$) or S2 invocation rate ($t{=}{-}0.28$, $p{=}0.78$). Under disturbance, return ($t{=}{-}0.78$, $p{=}0.44$) and S2 rate ($t{=}1.35$, $p{=}0.18$) remain insignificant. We adopt cold start for its simplicity. See \appdxref{apdx:ablation_warmstart} for full results.

\vspace{-4pt}

\paragraph{OOD distance metric.}
We compare Mahalanobis and Tanimoto distance \citep{rogers1960tanimoto} for OOD gating. A paired t-test across all task-config comparisons finds no significant difference in nominal return ($t{=}0.85$, $p{=}0.40$). In the \emph{theoretical} mode, Tanimoto yields equal or lower latency at every $\tau$ percentile, but degrades robustness: under perturbation, Mahalanobis achieves significantly higher return ($t{=}2.33$, $p{=}0.02$) by routing more timesteps to the planner. We therefore adopt Mahalanobis distance as the OOD gating. See \appdxref{apdx:ablation_tanimoto} for full results.

%% file: sections/limitations.tex
\section{Limitations and Discussions}
\label{sec:limitations}

\vspace{-4pt}

We identify three main limitations. First, the OOD gate is \emph{reactive}. It detects atypical states only after they are visited, rather than anticipating them from the learned dynamics model. A predictive gate could pre-route to System~2 before state degradation occurs, further reducing wasted computation on unrecoverable trajectories. Second, while the default $\tau$ is set by a fixed rule on held-out OOD scores, the optimal threshold varies across domains (e.g., locomotion vs.\ manipulation); real-world deployment may require tracing the Pareto frontier to identify a task-appropriate operating point. Finally, all experiments are conducted in simulation, while sim-to-real gaps exist for realistic robotic systems; we defer real-robot validation to future work.

%% file: sections/related_work.tex
\section{Related Work}
\label{sec:related_work}

\vspace{-4pt}

\paragraph{Data-Driven MPC and TD-MPC2.}
TD-MPC2 \citep{hansen2024tdmpc2} combines latent dynamics modeling with \gls{mppi} planning and \gls{td}-learning, achieving state-of-the-art performance across 104 continuous control tasks, outperforming SAC \citep{haarnoja2018sac} and DreamerV3 \citep{hafner2023dreamerv3}. Its ecosystem addresses training efficiency \citep{evers2026efficienttdmpc}, humanoid locomotion \citep{nguyen2025tdgrpc,nguyen2025doublyaware}, and world model reliability \citep{hansen2026hallucination}. Fast-TD-MPC builds on this framework but targets \emph{inference-time} acceleration.

\vspace{-4pt}

\paragraph{Accelerating Test-Time Planning.}
Approaches to reducing the per-step cost of \gls{mppi} fall into four categories: (i) \emph{Model reduction} distills the planner into a smaller or quantized model, but \gls{mppi} still runs at every step \citep{kuzmenko2025tdmpc_opt,wang2025bmpc}; (ii) \emph{Planner acceleration} speeds up the planning loop itself (e.g., parallelizing latent rollouts \citep{gao2026fastlewm}) but does not eliminate it; (iii) \emph{Full amortization} replaces the planner entirely with a feed-forward policy \citep{nguyen2026latentgeometry}, achieving large speedups but providing no fallback when the policy encounters out-of-distribution states; (iv) \emph{Conditional replanning} skips planning on some timesteps: \citet{cheng2026adarep} adapt replanning tolerance based on model mismatch by reusing cached plans, while \citet{lin2025speculation} generate action queues from the planner's own latent rollouts with full replanning on mismatch. Fast-TD-MPC differs by combining \emph{separately amortized} full policy replacement with OOD-gated fallback to the complete planner. The amortized policy is distilled from expert planner rollouts, so the planning expertise is retained in the fast path, and only the iterative optimization at inference time is eliminated. Classical event-triggered control \citep{tabuada2007event,heemels2012event} and the dual-process framework \citep{kahneman2011thinking,daw2005uncertainty,botvinick2009planning} provide broader intellectual context. See \appdxref{apdx:related_work} for a detailed comparison.

%% file: sections/conclusion.tex
\section{Conclusion}
\label{sec:conclusion}

\vspace{-4pt}

We presented Fast-TD-MPC, a framework that accelerates TD-MPC2 inference by routing in-distribution states to a compact amortized policy and falling back to the full \gls{mppi} planner only when the OOD detector flags atypical states. 
Across 103 continuous control tasks spanning four domains, Fast-TD-MPC achieves up to ${\sim}4\times$ inference speedup (${\sim}2.7\times$ on average) with ${<}10\%$ performance drop on 3 of 4 domains, using a single set of hyperparameters.
Under five combined external disturbance types, the gate adaptively increases planning on perturbed states, retaining 95\% of TD-MPC2's perturbed return on DMControl at half its latency. The framework is conceptually general: any model-based planner can serve as System 2, any amortized policy as System 1, and the gating decision reduces to a single transparent scalar comparison, a desirable property for safety-critical applications where
auditability matters.

%% file: appendix/appendix.tex
\noindent\makebox[\linewidth]{\rule{\linewidth}{0.4pt}}
\begin{center}
{\Large\bfseries APPENDICES}
\end{center}
\noindent\makebox[\linewidth]{\rule{\linewidth}{0.4pt}}

\vspace{1em}

\begin{list}{}{
    \setlength{\labelwidth}{2em}
    \setlength{\leftmargin}{2.5em}
    \setlength{\labelsep}{0.5em}
    \setlength{\rightmargin}{0pt}
}
\item \textbf{A.} \hyperref[apdx:method_details]{Method Details} \hfill \pageref{apdx:method_details}
\item \textbf{B.} \hyperref[apdx:task_visualizations]{Task Visualizations} \hfill \pageref{apdx:task_visualizations}
\item \textbf{C.} \hyperref[apdx:task_domains]{Task Domains} \hfill \pageref{apdx:task_domains}
\item \textbf{D.} \hyperref[apdx:env_details]{Environment Details} \hfill \pageref{apdx:env_details}
\item \textbf{E.} \hyperref[apdx:env_details_perturb]{Environment with Externally Applied Disturbances} \hfill \pageref{apdx:env_details_perturb}
\item \textbf{F.} \hyperref[apdx:full_results]{Full Results} \hfill \pageref{apdx:full_results}
\item \textbf{G.} \hyperref[apdx:full_perturbation_results]{Full Results under Disturbances} \hfill \pageref{apdx:full_perturbation_results}
\item \textbf{H.} \hyperref[apdx:evaluation_setup_and_hardware]{Evaluation Setup and Hardware Details} \hfill \pageref{apdx:evaluation_setup_and_hardware}
\item \textbf{I.} \hyperref[apdx:ablation_warmstart]{MPPI Planner Warm-Start Strategy Ablation} \hfill \pageref{apdx:ablation_warmstart}
\item \textbf{J.} \hyperref[apdx:ablation_tanimoto]{OOD Distance Metric Ablation} \hfill \pageref{apdx:ablation_tanimoto}
\item \textbf{K.} \hyperref[apdx:implementation_details]{Implementation Details} \hfill \pageref{apdx:implementation_details}
\item \textbf{L.} \hyperref[apdx:related_work]{Extended Related Work} \hfill \pageref{apdx:related_work}
\end{list}
\noindent\makebox[\linewidth]{\rule{\linewidth}{0.4pt}}
\vspace{10em}

\input{appendix/algorithm}
\clearpage

\input{appendix/task_visualizations}

\input{appendix/task_domains}

% \clearpage
\input{appendix/env_details}
\input{appendix/env_details_perturb}
\input{appendix/results_main_full}

% \clearpage
\input{appendix/results_main_full_perturbed}
\input{appendix/hardware_and_evaluation_setup}

\input{appendix/mppi_warmstart_ablation}
\input{appendix/ood_gating_ablation}

\input{appendix/implementation_details}
\input{appendix/extended_related_work}

%% file: appendix/algorithm.tex
\section{Method Details}
\label{apdx:method_details}

Algorithm~\ref{alg:fast_tdmpc} summarizes the full Fast-TD-MPC inference procedure. At each decision step $t$, the frozen encoder $\mathcal{E}_{\theta}$ maps the current state $s_t$ to latent $z_t$, and the OOD detector computes the Mahalanobis distance $g(z_t)$. If $g(z_t) \leq \tau$, the state is deemed in-distribution and the amortized policy $\pi_{\text{S1}}$ produces the action in a single forward pass. Otherwise, the full \gls{mppi} planner $\pi_{\text{S2}}$ is invoked. All components are frozen at deployment; the only runtime computation is the encoding, OOD scoring, and the selected policy's forward pass.

\begin{algorithm}[h]
\caption{Fast-TD-MPC inference}
\label{alg:fast_tdmpc}
\begin{algorithmic}[1]
\Require Frozen encoder $\mathcal{E}_{\theta}$, amortized policy $\pi_{\text{S1}}(\cdot\,;\phi)$, \gls{mppi} planner $\pi_{\text{S2}}(\cdot\,;\theta)$, OOD detector $g(\cdot\,;\bar{z}, \Sigma^{-1})$, threshold $\tau$
\State $s_0 \sim p_0$
\For{$t = 0, 1, \dots, T_{\max}-1$}
    \State $z_t \gets \mathcal{E}_{\theta}(s_t)$ \Comment{Encode state}
    \State $g_t \gets g(z_t\,;\bar{z}, \Sigma^{-1})$ \Comment{OOD score}
    \If{$g_t \leq \tau$}
        \State $a_t \gets \pi_{\text{S1}}(z_t;\phi)$ \Comment{System 1: fast policy}
    \Else
        \State $a_t \gets \pi_{\text{S2}}(z_t;\theta)$ \Comment{System 2: MPPI planner}
    \EndIf
    \State Execute $a_t$, observe $s_{t+1}$
\EndFor
\end{algorithmic}
\end{algorithm}

%% file: appendix/task_visualizations.tex
\section{Task Visualizations}
\label{apdx:task_visualizations}

\begin{figure}[H]
    \centering
    \includegraphics[width=0.85\linewidth]{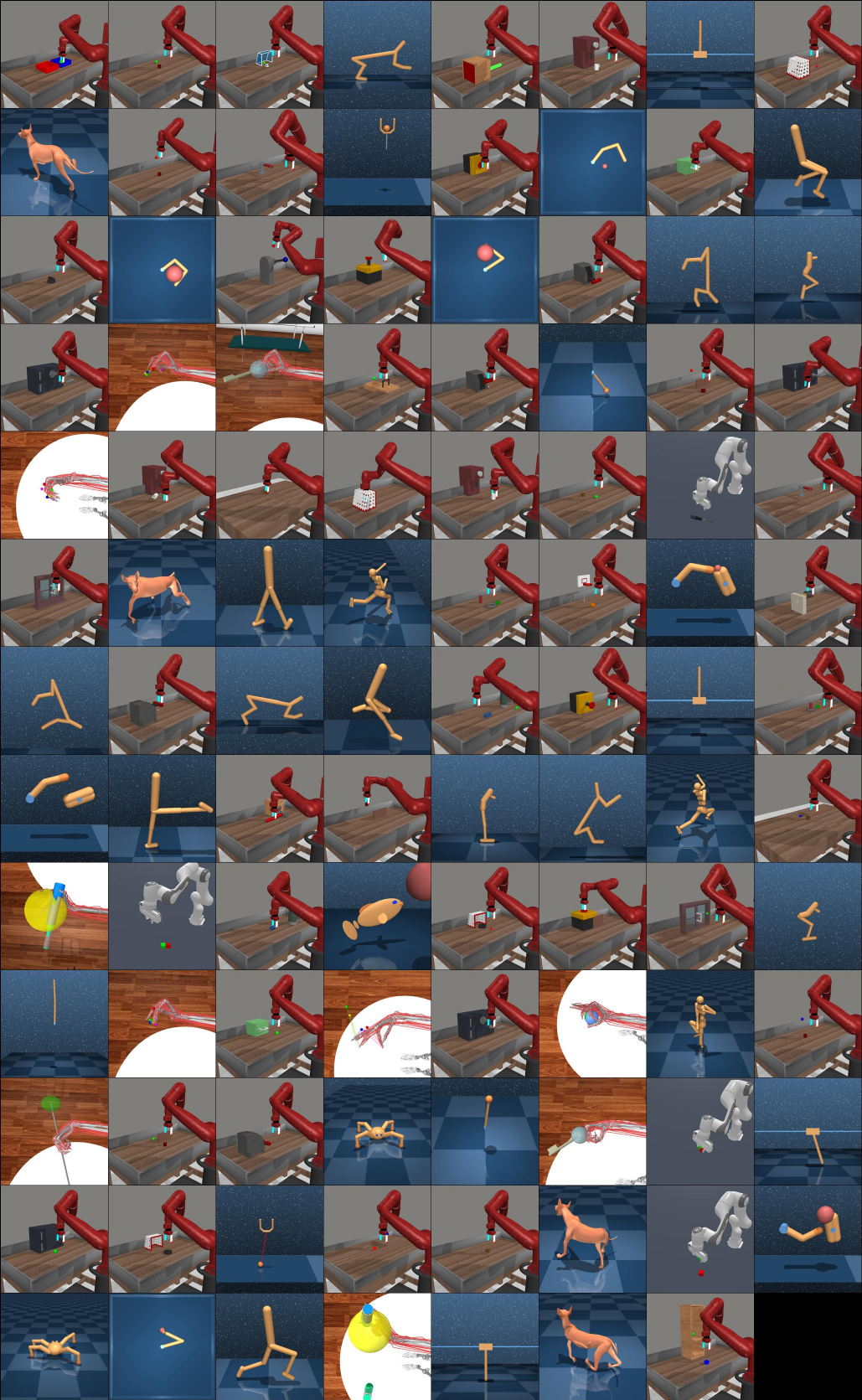}
    \caption{
    \textbf{Task visualizations.} Visualization of random states from all 103 diverse continuous control tasks used in our evaluation, spanning DMControl \citep{tassa2018dmcontrol}, Meta-World \citep{yu2019metaworld}, MyoSuite \citep{caggiano2022myosuite}, and ManiSkill2 \citep{gu2023maniskill2}.
    Tasks vary greatly in objective, embodiment, and action space. See \appdxref{apdx:task_domains} for task details.
    }
    \label{fig:task_grid}
\end{figure}

%% file: appendix/task_domains.tex
\section{Task Domains}
\label{apdx:task_domains}

We consider a total of 103 continuous control tasks from 4 task domains: DMControl \citep{tassa2018dmcontrol}, Meta-World \citep{yu2019metaworld}, MyoSuite \citep{caggiano2022myosuite}, and ManiSkill2 \citep{gu2023maniskill2}. We provide an exhaustive list of all tasks considered, along with their details in Table~\ref{tab:task_details_dmcontrol}, \ref{tab:task_details_metaworld}, \ref{tab:task_details_myosuite} and \ref{tab:task_details_maniskill2} below. Visualizations of each task are provided in \appdxref{apdx:task_visualizations}. Environment details are provided in \appdxref{apdx:env_details} and \appdxref{apdx:env_details_perturb}.

\begin{table}[h]
% \centering
\small
\setlength{\tabcolsep}{4.5pt}
\caption{\textbf{DMControl.} We consider a total 39 tasks from the DMControl domain. Observation dimension ranges from 3 - 223, with action dimensions ranging from 1 - 38. Metric: Return. Physics engine: MuJoCo.}
\label{tab:task_details_dmcontrol}
\begin{center}
\begin{tabular}{p{0.45\linewidth} r r}
\toprule
\textbf{Task} & \textbf{Obs.\ dim} & \textbf{Action dim} \\
\midrule
Acrobot Swingup & 6 & 1 \\
Cartpole Balance & 5 & 1 \\
Cartpole Balance Sparse & 5 & 1 \\
Cartpole Swingup & 5 & 1 \\
Cartpole Swingup Sparse & 5 & 1 \\
Cheetah Jump & 17 & 6 \\
Cheetah Run & 17 & 6 \\
Cheetah Run Back & 17 & 6 \\
Cheetah Run Backwards & 17 & 6 \\
Cheetah Run Front & 17 & 6 \\
Cup Catch & 8 & 2 \\
Cup Spin & 8 & 2 \\
Dog Run & 223 & 38 \\
Dog Stand & 223 & 38 \\
Dog Trot & 223 & 38 \\
Dog Walk & 223 & 38 \\
Finger Spin & 9 & 2 \\
Finger Turn Easy & 12 & 2 \\
Finger Turn Hard & 12 & 2 \\
Fish Swim & 24 & 5 \\
Hopper Hop & 15 & 4 \\
Hopper Hop Backwards & 15 & 4 \\
Hopper Stand & 15 & 4 \\
Humanoid Run & 67 & 21 \\
Humanoid Stand & 67 & 21 \\
Humanoid Walk & 67 & 21 \\
Pendulum Spin & 3 & 1 \\
Pendulum Swingup & 3 & 1 \\
Quadruped Run & 78 & 12 \\
Quadruped Walk & 78 & 12 \\
Reacher Easy & 6 & 2 \\
Reacher Hard & 6 & 2 \\
Reacher Three Easy & 8 & 3 \\
Reacher Three Hard & 8 & 3 \\
Walker Run & 24 & 6 \\
Walker Run Backwards & 24 & 6 \\
Walker Stand & 24 & 6 \\
Walker Walk & 24 & 6 \\
Walker Walk Backwards & 24 & 6 \\
\bottomrule
\end{tabular}
\end{center}
\end{table}

\begin{table}[h]
% \centering
\small
\setlength{\tabcolsep}{4.5pt}
\caption{\textbf{Meta-World.} We consider a total 50 tasks from the Meta-World domain. Observation dimensions are fixed at 39, with action dimensions fixed at 4. Metric: Success rate. Physics engine: MuJoCo.}
\label{tab:task_details_metaworld}
\begin{center}
\begin{tabular}{p{0.45\linewidth} r r}
\toprule
\textbf{Task} & \textbf{Obs.\ dim} & \textbf{Action dim} \\
\midrule
Assembly & 39 & 4 \\
Basketball & 39 & 4 \\
Bin Picking & 39 & 4 \\
Box Close & 39 & 4 \\
Button Press & 39 & 4 \\
Button Press Topdown & 39 & 4 \\
Button Press Topdown Wall & 39 & 4 \\
Button Press Wall & 39 & 4 \\
Coffee Button & 39 & 4 \\
Coffee Pull & 39 & 4 \\
Coffee Push & 39 & 4 \\
Dial Turn & 39 & 4 \\
Disassemble & 39 & 4 \\
Door Close & 39 & 4 \\
Door Lock & 39 & 4 \\
Door Open & 39 & 4 \\
Door Unlock & 39 & 4 \\
Drawer Close & 39 & 4 \\
Drawer Open & 39 & 4 \\
Faucet Close & 39 & 4 \\
Faucet Open & 39 & 4 \\
Hammer & 39 & 4 \\
% Hand Insert & 39 & 4 \\
% Handle Press & 39 & 4 \\
% Handle Press Side & 39 & 4 \\
% Handle Pull & 39 & 4 \\
% Handle Pull Side & 39 & 4 \\
% Lever Pull & 39 & 4 \\
% Peg Insert Side & 39 & 4 \\
% Peg Unplug Side & 39 & 4 \\
% Pick Out Of Hole & 39 & 4 \\
% Pick Place & 39 & 4 \\
% Pick Place Wall & 39 & 4 \\
% Plate Slide & 39 & 4 \\
% Plate Slide Back & 39 & 4 \\
% Plate Slide Back Side & 39 & 4 \\
% Plate Slide Side & 39 & 4 \\
% Push & 39 & 4 \\
% Push Back & 39 & 4 \\
% Push Wall & 39 & 4 \\
% Reach & 39 & 4 \\
% Reach Wall & 39 & 4 \\
% Shelf Place & 39 & 4 \\
% Soccer & 39 & 4 \\
% Stick Pull & 39 & 4 \\
% Stick Push & 39 & 4 \\
% Sweep & 39 & 4 \\
% Sweep Into & 39 & 4 \\
% Window Close & 39 & 4 \\
... & ... & ... \\
Window Open & 39 & 4 \\
\bottomrule
\end{tabular}
\end{center}
\end{table}

\begin{table}[H]
% \centering
\small
\setlength{\tabcolsep}{4.5pt}
\caption{\textbf{MyoSuite.} We consider a total 10 tasks from the MyoSuite domain. Observation dimensions ranges from 83 - 115, with action dimensions fixed at 39. Metric: Success rate. Physics engine: MuJoCo.}
\label{tab:task_details_myosuite}
\begin{center}
\begin{tabular}{p{0.45\linewidth} r r}
\toprule
\textbf{Task} & \textbf{Obs.\ dim} & \textbf{Action dim} \\
\midrule
Myo Hand Key Turn & 93 & 39 \\
Myo Hand Key Turn Hard & 93 & 39 \\
Myo Hand Obj Hold & 91 & 39 \\
Myo Hand Obj Hold Hard & 91 & 39 \\
Myo Hand Pen Twirl & 83 & 39 \\
Myo Hand Pen Twirl Hard & 83 & 39 \\
Myo Hand Pose & 108 & 39 \\
Myo Hand Pose Hard & 108 & 39 \\
Myo Hand Reach & 115 & 39 \\
Myo Hand Reach Hard & 115 & 39 \\
\bottomrule
\end{tabular}
\end{center}
\end{table}

\begin{table}[H]
% \centering
\small
\setlength{\tabcolsep}{4.5pt}
\caption{\textbf{ManiSkill2.} We consider a total 4 tasks from the ManiSkill2 domain. Observation dimensions ranges from 42 - 55, with action dimensions ranging from 4 - 7. Metric: Success rate. Physics engine: SAPIEN.}
\label{tab:task_details_maniskill2}
\begin{center}
\begin{tabular}{p{0.45\linewidth} r r}
\toprule
\textbf{Task} & \textbf{Obs.\ dim} & \textbf{Action dim} \\
\midrule
Lift Cube & 42 & 4 \\
Pick Cube & 51 & 4 \\
Pick Ycb & 51 & 7 \\
Stack Cube & 55 & 4 \\
\bottomrule
\end{tabular}
\end{center}
\end{table}

%% file: appendix/env_details.tex
\section{Environment Details}
\label{apdx:env_details}

For experiments on \textit{\textbf{RQ1}}, we follow \citet{hansen2024tdmpc2}'s setup and benchmark on DMControl, Meta-World, ManiSkill2, and MyoSuite without modification. All four domains are infinite-horizon continuous control environments with fixed episode lengths and no termination conditions. We list episode lengths, action repeats, and the performance metric used for each domain in Table~\ref{tab:env_details}. Following \citet{hansen2024tdmpc2}, we adopt a strict success criterion: an episode is successful only if its final step is successful. In object manipulation tasks, this ensures that an episode in which an object is picked up but subsequently dropped is not counted as a success.

\begin{table}[H]
% \centering
\caption{\textbf{Environment details.} We list the episode length and action repeat used for each task domain, as well as the total number of environment steps and performance metrics that we use for benchmarking methods. All methods use the same values for all tasks.}
\label{tab:env_details}
\begin{center}
\begin{tabular}{lcccc}
\toprule
& \textbf{DMControl} & \textbf{Meta-World} & \textbf{ManiSkill2} & \textbf{MyoSuite} \\
\midrule
Episode length     & 1,000 & 200 & 200 & 100 \\
Action repeat      & 2     & 2   & 2   & 1   \\
Effective length   & 500   & 100 & 100 & 100 \\
Performance metric & Reward & Success & Success & Success \\
\bottomrule
\end{tabular}
\end{center}
\end{table}

%% file: appendix/env_details_perturb.tex
\section{Environment with Externally Applied Disturbances}
\label{apdx:env_details_perturb}

For experiments on \textit{\textbf{RQ2}}, we introduce five types of disturbances to all four benchmark domains. Table~\ref{tab:perturbation-params} details the five disturbance types, organized following the taxonomy of \citet{gu2025robustgym} into observation, action, and environment disruptors, along with their categories and activation probabilities. All disturbances are applied post-training at evaluation time with per-episode stochasticity (seed\,=\,42); per-episode parameters are drawn from the specified ranges using an isolated \texttt{RandomState}.

\begin{table}[H]
\small
\caption{\textbf{Environment disturbance configurations.} $p$ denotes the per-step activation probability. $^{\dagger}$1--2 kicks per episode at random timesteps within 10--90\% of episode length. Obs.\ noise and vel.\ kick magnitudes are relative (scaled by per-episode std). Tasks: 102 (39 DMControl $+$ 50 MetaWorld $+$ 10 MyoSuite $+$ 3 ManiSkill2; \texttt{pick-ycb} excluded). 
}
\label{tab:perturbation-params}
\begin{center}
\begin{tabular}{llcll}
\toprule
\textbf{Disturbance} & \textbf{Cat.} & $p$ & \textbf{Mechanism} & \textbf{Parameter} \\
\midrule
Obs.\ noise & Obs.\ & 1.0 & Relative $\sigma\!\cdot\!\text{std}(s)$ on $s_t$ & $\sigma = .005$ \\
Act.\ noise & Act.\ & 1.0 & Additive $\mathcal{N}(0, \sigma^2)$ on $a_t$, clamp $[-1,1]$ & $\sigma = .005$ \\
Vel.\ kick & Env.\ & 1.0$^{\dagger}$ & Impulse on $q_{\text{vel}}$ at 1--2 random $t$ & $m_{\max} = .3$ \\
Grav.\ shift & Env.\ & 1.0 & Sustained $g \leftarrow g \cdot k$ per episode & $\delta = .05$ \\
Sus.\ force & Env.\ & 0.15 & Per-step force injection when active & $f_{\max} = .3$ \\
\midrule
Combined & All & --- & All five applied simultaneously & --- \\
\bottomrule
\end{tabular}
\end{center}
\end{table}

\paragraph{Excluded Task.}
The \texttt{pick-ycb} task (PickSingleYCB-v0) from ManiSkill2 is excluded from the perturbation evaluation due to a SAPIEN \citep{xiang2020sapien} memory management bug that triggers a C-level segfault when physics-based disturbances are applied to articulated YCB mesh objects. This is a known SAPIEN limitation unrelated to our method. All other ManiSkill2 tasks are unaffected.

\paragraph{SAPIEN Backend Adaptation.}
ManiSkill2 environments are built on SAPIEN \citep{xiang2020sapien}, which exposes a different physics API than MuJoCo \citep{todorov2012mujoco}. While observation noise and action noise operate on the observation and action arrays respectively (backend-agnostic), three perturbation types require adaptation: (i)~\emph{Gravity shift:} SAPIEN's \texttt{SceneConfig.gravity} is a direct 3D vector, equivalent to MuJoCo's \texttt{model.opt.gravity}. (ii)~\emph{Velocity kick:} applied via \texttt{set\_qvel(get\_qvel() + impulse)}. (iii)~\emph{Sustained force:} SAPIEN does not expose \texttt{qfrc\_applied}; we implement this as per-step velocity injection, \texttt{set\_qvel(get\_qvel() + $\mathbf{f}$)}, producing equivalent trajectory disruption with the same magnitude and direction sampling.

%% file: appendix/results_main_full.tex
\section{Full Results}
\label{apdx:full_results}

Tables~\ref{tab:appendix_main_dmcontrol} -- \ref{tab:appendix_main_maniskill2} present the full benchmark results for each domain, including per-step latency, S2 invocation rate ($\rho$), and peak GPU memory, complementing Table~\ref{tab:main_results}. Tables~\ref{tab:per_task_clean_dmcontrol} -- \ref{tab:per_task_clean_maniskill2} provide per-task breakdowns for four key methods.

\begin{table}[h]
% \centering
% \scriptsize
\small
\setlength{\tabcolsep}{4.5pt}
\providecommand{\pmci}[2]{$#1_{\scriptscriptstyle \pm #2}$}
\caption{\textbf{DMControl full results.} 39 tasks, 3 seeds each. Mean and 95\% CI across all tasks. Return, per-step latency, S2 invocation rate ($\rho$), and peak GPU memory. \textsuperscript{\dag}Referenced from \citet{hansen2024tdmpc2}. \emph{all}: theoretical OOD mode; \emph{RG}: reward-gated mode.}
\label{tab:appendix_main_dmcontrol}
\begin{center}
\begin{tabular}{l r r r r}
\toprule
\textbf{Method} & \textbf{Return} & \textbf{Lat.}~{(ms)} & $\rho$~{(\%)} & \textbf{GPU}~{(MB)} \\
\midrule
DreamerV3\textsuperscript{$\dag$} & \pmci{735}{93} & \pmci{3.2}{.1} & --- & --- \\
SAC\textsuperscript{$\dag$} & \pmci{631}{108} & \pmci{2.0}{.4} & --- & --- \\
\midrule
S2-alone (TD-MPC2) & \pmci{820}{55} & \pmci{27.3}{1.1} & $100$ & \pmci{95}{0} \\
S2-alone ($N{=}64$) & \pmci{639}{100} & \pmci{26.9}{1.1} & $100$ & \pmci{69}{0} \\
S2 policy-only & \pmci{560}{105} & \pmci{1.9}{.3} & $0$ & \pmci{66}{0} \\
S1 alone & \pmci{623}{91} & \pmci{1.6}{.2} & $0$ & \pmci{68}{0} \\
\midrule
Round-robin (S2 50\%) & \pmci{813}{57} & \pmci{14.4}{.7} & $50$ & \pmci{96}{0} \\
Round-robin (S2 33\%) & \pmci{782}{66} & \pmci{10.1}{.5} & $33$ & \pmci{68}{0} \\
Round-robin (S2 20\%) & \pmci{743}{74} & \pmci{6.6}{.4} & $20$ & \pmci{96}{0} \\
Round-robin (S2 10\%) & \pmci{699}{83} & \pmci{4.0}{.3} & $10$ & \pmci{96}{0} \\
\midrule
Fast-TD-MPC (\emph{all}, $\tau_{p50}$) & \pmci{829}{55} & \pmci{19.4}{2.4} & \pmci{68}{7} & \pmci{85}{3} \\
Fast-TD-MPC (\emph{all}, $\tau_{p75}$) & \pmci{823}{55} & \pmci{12.2}{2.7} & \pmci{40}{9} & \pmci{77}{2} \\
Fast-TD-MPC (\emph{all}, $\tau_{p90}$) & \pmci{785}{58} & \pmci{6.5}{2.0} & \pmci{18}{7} & \pmci{73}{2} \\
Fast-TD-MPC (\emph{all}, $\tau_{p95}$) & \pmci{745}{74} & \pmci{3.8}{1.4} & \pmci{8}{4} & \pmci{70}{1} \\
Fast-TD-MPC (\emph{all}, $\tau_{p99}$) & \pmci{650}{92} & \pmci{1.9}{.3} & \pmci{1}{0} & \pmci{69}{0} \\
\midrule
Fast-TD-MPC (\emph{RG}, $\tau_{p25}$) & \pmci{784}{68} & \pmci{10.8}{2.3} & \pmci{35}{8} & \pmci{77}{2} \\
Fast-TD-MPC (\emph{RG}, $\tau_{p50}$) & \pmci{764}{70} & \pmci{8.1}{1.8} & \pmci{25}{7} & \pmci{74}{2} \\
Fast-TD-MPC (\emph{RG}, $\tau_{p75}$) & \pmci{744}{74} & \pmci{6.4}{1.5} & \pmci{18}{5} & \pmci{72}{2} \\
\bottomrule
\end{tabular}
\end{center}
\end{table}

\begin{table}[h]
% \centering
% \scriptsize
\small
\setlength{\tabcolsep}{4.5pt}
\providecommand{\pmci}[2]{$#1_{\scriptscriptstyle \pm #2}$}
\caption{\textbf{Meta-World full results.} 50 tasks, 3 seeds each. Mean and 95\% CI across all tasks. Success rate, per-step latency, S2 invocation rate ($\rho$), and peak GPU memory. \textsuperscript{\dag}Referenced from \citet{hansen2024tdmpc2}. \emph{all}: theoretical OOD mode; \emph{RG}: reward-gated mode.}
\label{tab:appendix_main_metaworld}
\begin{center}
\begin{tabular}{l r r r r}
\toprule
\textbf{Method} & \textbf{Succ.} & \textbf{Lat.}~{(ms)} & $\rho$~{(\%)} & \textbf{GPU}~{(MB)} \\
\midrule
DreamerV3\textsuperscript{$\dag$} & \pmci{70}{8} & \pmci{3.8}{.1} & --- & --- \\
SAC\textsuperscript{$\dag$} & \pmci{84}{7} & \pmci{2.0}{.2} & --- & --- \\
\midrule
S2-alone (TD-MPC2) & \pmci{84}{9} & \pmci{26.9}{.2} & $100$ & \pmci{95}{0} \\
S2-alone ($N{=}64$) & \pmci{85}{7} & \pmci{26.7}{.2} & $100$ & \pmci{69}{0} \\
S2 policy-only & \pmci{82}{8} & \pmci{2.7}{.2} & $0$ & \pmci{66}{0} \\
S1 alone & \pmci{87}{8} & \pmci{2.5}{.2} & $0$ & \pmci{67}{0} \\
\midrule
Round-robin (S2 50\%) & \pmci{86}{7} & \pmci{14.7}{.2} & $50$ & \pmci{96}{0} \\
Round-robin (S2 33\%) & \pmci{86}{8} & \pmci{10.7}{.2} & $33$ & \pmci{67}{0} \\
Round-robin (S2 20\%) & \pmci{85}{8} & \pmci{7.3}{.2} & $20$ & \pmci{96}{0} \\
Round-robin (S2 10\%) & \pmci{86}{8} & \pmci{4.9}{.2} & $10$ & \pmci{96}{0} \\
\midrule
Fast-TD-MPC (\emph{all}, $\tau_{p50}$) & \pmci{85}{8} & \pmci{24.8}{1.1} & \pmci{90}{4} & \pmci{95}{1} \\
Fast-TD-MPC (\emph{all}, $\tau_{p75}$) & \pmci{85}{8} & \pmci{18.7}{1.5} & \pmci{66}{6} & \pmci{92}{2} \\
Fast-TD-MPC (\emph{all}, $\tau_{p90}$) & \pmci{85}{8} & \pmci{12.2}{1.6} & \pmci{39}{6} & \pmci{90}{2} \\
Fast-TD-MPC (\emph{all}, $\tau_{p95}$) & \pmci{85}{8} & \pmci{9.8}{1.5} & \pmci{29}{6} & \pmci{88}{2} \\
Fast-TD-MPC (\emph{all}, $\tau_{p99}$) & \pmci{85}{8} & \pmci{6.2}{1.3} & \pmci{15}{5} & \pmci{85}{2} \\
\midrule
Fast-TD-MPC (\emph{RG}, $\tau_{p25}$) & \pmci{85}{8} & \pmci{12.3}{1.7} & \pmci{40}{7} & \pmci{85}{2} \\
Fast-TD-MPC (\emph{RG}, $\tau_{p50}$) & \pmci{85}{8} & \pmci{10.0}{1.5} & \pmci{30}{6} & \pmci{80}{2} \\
Fast-TD-MPC (\emph{RG}, $\tau_{p75}$) & \pmci{84}{8} & \pmci{8.6}{1.5} & \pmci{25}{6} & \pmci{79}{2} \\
\bottomrule
\end{tabular}
\end{center}
\end{table}

\begin{table}[h]
% \centering
% \scriptsize
\small
\setlength{\tabcolsep}{4.5pt}
\providecommand{\pmci}[2]{$#1_{\scriptscriptstyle \pm #2}$}
\caption{\textbf{MyoSuite full results.} 10 tasks, 3 seeds each. Mean and 95\% CI across all tasks. Success rate, per-step latency, S2 invocation rate ($\rho$), and peak GPU memory. \textsuperscript{\dag}Referenced from \citet{hansen2024tdmpc2}. \emph{all}: theoretical OOD mode; \emph{RG}: reward-gated mode.}
\label{tab:appendix_main_myosuite}
\begin{center}
\begin{tabular}{l r r r r}
\toprule
\textbf{Method} & \textbf{Succ.} & \textbf{Lat.}~{(ms)} & $\rho$~{(\%)} & \textbf{GPU}~{(MB)} \\
\midrule
DreamerV3\textsuperscript{$\dag$} & \pmci{14}{17} & \pmci{4.2}{.1} & --- & --- \\
SAC\textsuperscript{$\dag$} & \pmci{38}{20} & \pmci{2.9}{.5} & --- & --- \\
\midrule
S2-alone (TD-MPC2) & \pmci{40}{20} & \pmci{35.5}{.4} & $100$ & \pmci{97}{0} \\
S2-alone ($N{=}64$) & \pmci{24}{16} & \pmci{35.4}{.4} & $100$ & \pmci{70}{0} \\
S2 policy-only & \pmci{18}{15} & \pmci{3.8}{.4} & $0$ & \pmci{67}{0} \\
S1 alone & \pmci{35}{16} & \pmci{3.6}{.3} & $0$ & \pmci{69}{0} \\
\midrule
Round-robin (S2 50\%) & \pmci{45}{21} & \pmci{19.3}{.3} & $50$ & \pmci{98}{0} \\
Round-robin (S2 33\%) & \pmci{41}{18} & \pmci{14.1}{.3} & $33$ & \pmci{68}{0} \\
Round-robin (S2 20\%) & \pmci{40}{22} & \pmci{9.8}{.3} & $20$ & \pmci{98}{0} \\
Round-robin (S2 10\%) & \pmci{40}{21} & \pmci{6.6}{.3} & $10$ & \pmci{98}{0} \\
\midrule
Fast-TD-MPC (\emph{all}, $\tau_{p50}$) & \pmci{39}{21} & \pmci{31.4}{3.6} & \pmci{88}{12} & \pmci{91}{5} \\
Fast-TD-MPC (\emph{all}, $\tau_{p75}$) & \pmci{37}{21} & \pmci{26.6}{5.8} & \pmci{72}{18} & \pmci{93}{4} \\
Fast-TD-MPC (\emph{all}, $\tau_{p90}$) & \pmci{37}{21} & \pmci{22.6}{7.6} & \pmci{60}{24} & \pmci{85}{5} \\
Fast-TD-MPC (\emph{all}, $\tau_{p95}$) & \pmci{38}{23} & \pmci{19.8}{7.6} & \pmci{51}{24} & \pmci{82}{5} \\
Fast-TD-MPC (\emph{all}, $\tau_{p99}$) & \pmci{28}{17} & \pmci{10.9}{5.8} & \pmci{23}{19} & \pmci{78}{5} \\
\midrule
Fast-TD-MPC (\emph{RG}, $\tau_{p25}$) & \pmci{39}{22} & \pmci{24.5}{4.6} & \pmci{66}{15} & \pmci{89}{5} \\
Fast-TD-MPC (\emph{RG}, $\tau_{p50}$) & \pmci{35}{21} & \pmci{20.6}{5.2} & \pmci{54}{17} & \pmci{85}{5} \\
Fast-TD-MPC (\emph{RG}, $\tau_{p75}$) & \pmci{39}{21} & \pmci{15.7}{4.7} & \pmci{39}{15} & \pmci{83}{5} \\
\bottomrule
\end{tabular}
\end{center}
\end{table}

\begin{table}[h]
% \centering
% \scriptsize
\small
\setlength{\tabcolsep}{4.5pt}
\providecommand{\pmci}[2]{$#1_{\scriptscriptstyle \pm #2}$}
\caption{\textbf{ManiSkill2 full results.} 4 tasks, 3 seeds each. Mean and 95\% CI across all tasks. Success rate, per-step latency, S2 invocation rate ($\rho$), and peak GPU memory. \textsuperscript{\dag}Referenced from \citet{hansen2024tdmpc2}. \emph{all}: theoretical OOD mode; \emph{RG}: reward-gated mode.}
\label{tab:appendix_main_maniskill2}
\begin{center}
\begin{tabular}{l r r r r}
\toprule
\textbf{Method} & \textbf{Succ.} & \textbf{Lat.}~{(ms)} & $\rho$~{(\%)} & \textbf{GPU}~{(MB)} \\
\midrule
DreamerV3\textsuperscript{$\dag$} & \pmci{34}{56} & \pmci{8.7}{.7} & --- & --- \\
SAC\textsuperscript{$\dag$} & \pmci{22}{54} & \pmci{5.8}{.7} & --- & --- \\
\midrule
S2-alone (TD-MPC2) & \pmci{65}{54} & \pmci{45.9}{4.8} & $100$ & \pmci{70}{0} \\
S2-alone ($N{=}64$) & \pmci{63}{53} & \pmci{41.5}{3.4} & $100$ & \pmci{44}{0} \\
S2 policy-only & \pmci{48}{56} & \pmci{11.1}{2.5} & $0$ & \pmci{41}{0} \\
S1 alone & \pmci{19}{39} & \pmci{10.7}{3.9} & $0$ & \pmci{42}{0} \\
\midrule
Round-robin (S2 50\%) & \pmci{60}{44} & \pmci{28.4}{5.6} & $50$ & \pmci{72}{0} \\
Round-robin (S2 33\%) & \pmci{42}{50} & \pmci{22.3}{3.8} & $33$ & \pmci{42}{0} \\
Round-robin (S2 20\%) & \pmci{38}{48} & \pmci{17.6}{3.9} & $20$ & \pmci{72}{0} \\
Round-robin (S2 10\%) & \pmci{28}{46} & \pmci{14.5}{4.7} & $10$ & \pmci{72}{0} \\
\midrule
Fast-TD-MPC (\emph{all}, $\tau_{p50}$) & \pmci{54}{55} & \pmci{33.9}{12.1} & \pmci{67}{32} & \pmci{63}{8} \\
Fast-TD-MPC (\emph{all}, $\tau_{p75}$) & \pmci{47}{54} & \pmci{23.3}{10.7} & \pmci{34}{26} & \pmci{46}{5} \\
Fast-TD-MPC (\emph{all}, $\tau_{p90}$) & \pmci{32}{42} & \pmci{16.0}{5.2} & \pmci{13}{9} & \pmci{46}{5} \\
Fast-TD-MPC (\emph{all}, $\tau_{p95}$) & \pmci{28}{46} & \pmci{14.2}{3.9} & \pmci{8}{9} & \pmci{46}{5} \\
Fast-TD-MPC (\emph{all}, $\tau_{p99}$) & \pmci{20}{39} & \pmci{11.5}{3.9} & \pmci{1}{2} & \pmci{46}{5} \\
\midrule
Fast-TD-MPC (\emph{RG}, $\tau_{p25}$) & \pmci{48}{43} & \pmci{28.5}{20.9} & \pmci{51}{65} & \pmci{58}{9} \\
Fast-TD-MPC (\emph{RG}, $\tau_{p50}$) & \pmci{43}{43} & \pmci{25.6}{18.9} & \pmci{40}{55} & \pmci{51}{8} \\
Fast-TD-MPC (\emph{RG}, $\tau_{p75}$) & \pmci{39}{51} & \pmci{20.9}{10.8} & \pmci{25}{29} & \pmci{46}{5} \\
\bottomrule
\end{tabular}
\end{center}
\end{table}

\begin{table}[h]
% \centering
\scriptsize
\setlength{\tabcolsep}{3.5pt}
\providecommand{\pmci}[2]{$#1_{\scriptscriptstyle \pm #2}$}
\caption{\textbf{DMControl per-task results.} 39 tasks, 3 seeds each. Mean and 95\% CI per task. Return and per-step latency (ms) for 4 key methods. \emph{all}: theoretical OOD mode.}
\label{tab:per_task_clean_dmcontrol}
\begin{center}
\resizebox{\linewidth}{!}{%
\begin{tabular}{l r r r r r r r r}
\toprule
\textbf{Task} & \multicolumn{2}{c}{S2-alone (TD-MPC2)} & \multicolumn{2}{c}{S1 alone} & \multicolumn{2}{c}{Fast-TD-MPC (all, $\tau_{p50}$)} & \multicolumn{2}{c}{Fast-TD-MPC (all, $\tau_{p90}$)} \\
\cmidrule(lr){2-3} \cmidrule(lr){4-5} \cmidrule(lr){6-7} \cmidrule(lr){8-9}
 & \textbf{Return} & \textbf{Lat.}~{(ms)} & \textbf{Return} & \textbf{Lat.}~{(ms)} & \textbf{Return} & \textbf{Lat.}~{(ms)} & \textbf{Return} & \textbf{Lat.}~{(ms)} \\
\midrule
Acrobot Swingup & \pmci{642}{25} & \pmci{26.2}{.8} & \pmci{523}{163} & \pmci{1.2}{.0} & \pmci{638}{87} & \pmci{9.8}{3.1} & \pmci{582}{155} & \pmci{2.6}{.9} \\
Cartpole Balance Sparse & $1000$ & \pmci{26.0}{1.7} & $1000$ & \pmci{1.3}{.0} & $1000$ & \pmci{14.8}{23.9} & $1000$ & \pmci{4.8}{5.0} \\
Cartpole Balance & \pmci{999}{1} & \pmci{25.8}{1.4} & \pmci{1000}{1} & \pmci{1.4}{.0} & \pmci{999}{1} & \pmci{21.1}{11.7} & \pmci{999}{1} & \pmci{3.4}{1.2} \\
Cartpole Swingup Sparse & \pmci{297}{1189} & \pmci{25.6}{1.3} & \pmci{284}{1216} & \pmci{1.2}{.0} & \pmci{284}{1216} & \pmci{17.3}{13.9} & \pmci{303}{1177} & \pmci{3.8}{1.1} \\
Cartpole Swingup & \pmci{661}{954} & \pmci{26.2}{1.1} & \pmci{665}{938} & \pmci{1.4}{.1} & \pmci{654}{987} & \pmci{14.4}{4.1} & \pmci{666}{934} & \pmci{3.5}{1.5} \\
Cheetah Jump & \pmci{833}{12} & \pmci{26.4}{1.1} & \pmci{779}{122} & \pmci{1.3}{.2} & \pmci{833}{6} & \pmci{12.4}{1.4} & \pmci{828}{9} & \pmci{3.4}{2.2} \\
\bottomrule
\end{tabular}}
\end{center}
\end{table}

\begin{table}[h]
% \centering
\scriptsize
\setlength{\tabcolsep}{3.5pt}
\providecommand{\pmci}[2]{$#1_{\scriptscriptstyle \pm #2}$}
\caption{\textbf{Meta-World per-task results.} 50 tasks, 3 seeds each. Mean and 95\% CI per task. Success rate (\%) and per-step latency (ms) for 4 key methods. \emph{all}: theoretical OOD mode.}
\label{tab:per_task_clean_metaworld}
\begin{center}
\resizebox{\linewidth}{!}{%
\begin{tabular}{l r r r r r r r r}
\toprule
\textbf{Task} & \multicolumn{2}{c}{S2-alone (TD-MPC2)} & \multicolumn{2}{c}{S1 alone} & \multicolumn{2}{c}{Fast-TD-MPC (all, $\tau_{p50}$)} & \multicolumn{2}{c}{Fast-TD-MPC (all, $\tau_{p90}$)} \\
\cmidrule(lr){2-3} \cmidrule(lr){4-5} \cmidrule(lr){6-7} \cmidrule(lr){8-9}
 & \textbf{Succ.} & \textbf{Lat.}~{(ms)} & \textbf{Succ.} & \textbf{Lat.}~{(ms)} & \textbf{Succ.} & \textbf{Lat.}~{(ms)} & \textbf{Succ.} & \textbf{Lat.}~{(ms)} \\
\midrule
Assembly & $100$ & \pmci{26.4}{1.1} & \pmci{67}{143} & \pmci{2.2}{.2} & \pmci{90}{25} & \pmci{27.1}{1.6} & \pmci{87}{57} & \pmci{19.7}{24.9} \\
Basketball & \pmci{97}{14} & \pmci{26.2}{.9} & $100$ & \pmci{2.0}{.1} & \pmci{90}{43} & \pmci{26.7}{.1} & \pmci{83}{72} & \pmci{20.6}{12.0} \\
Bin Picking & $100$ & \pmci{26.8}{1.4} & $100$ & \pmci{2.1}{.2} & \pmci{97}{14} & \pmci{27.1}{2.3} & $100$ & \pmci{14.5}{15.7} \\
Box Close & $100$ & \pmci{27.6}{.5} & $100$ & \pmci{3.4}{.6} & $100$ & \pmci{27.8}{1.2} & $100$ & \pmci{9.7}{7.4} \\
Button Press Topdown Wall & $100$ & \pmci{26.9}{.7} & $100$ & \pmci{2.7}{.2} & $100$ & \pmci{17.0}{9.5} & $100$ & \pmci{5.9}{2.0} \\
Button Press Topdown & $100$ & \pmci{26.9}{2.0} & $100$ & \pmci{2.7}{.3} & $100$ & \pmci{15.6}{6.1} & $100$ & \pmci{6.3}{1.2} \\
\bottomrule
\end{tabular}}
\end{center}
\end{table}

\begin{table}[h]
% \centering
\scriptsize
\setlength{\tabcolsep}{3.5pt}
\providecommand{\pmci}[2]{$#1_{\scriptscriptstyle \pm #2}$}
\caption{\textbf{MyoSuite per-task results.} 10 tasks, 3 seeds each. Mean and 95\% CI per task. Success rate (\%) and per-step latency (ms) for 4 key methods. \emph{all}: theoretical OOD mode.}
\label{tab:per_task_clean_myosuite}
\begin{center}
\resizebox{\linewidth}{!}{%
\begin{tabular}{l r r r r r r r r}
\toprule
\textbf{Task} & \multicolumn{2}{c}{S2-alone (TD-MPC2)} & \multicolumn{2}{c}{S1 alone} & \multicolumn{2}{c}{Fast-TD-MPC (all, $\tau_{p50}$)} & \multicolumn{2}{c}{Fast-TD-MPC (all, $\tau_{p90}$)} \\
\cmidrule(lr){2-3} \cmidrule(lr){4-5} \cmidrule(lr){6-7} \cmidrule(lr){8-9}
 & \textbf{Succ.} & \textbf{Lat.}~{(ms)} & \textbf{Succ.} & \textbf{Lat.}~{(ms)} & \textbf{Succ.} & \textbf{Lat.}~{(ms)} & \textbf{Succ.} & \textbf{Lat.}~{(ms)} \\
\midrule
MyoHand Key Turn Hard & \pmci{30}{66} & \pmci{36.6}{1.5} & \pmci{20}{86} & \pmci{4.1}{.5} & \pmci{27}{57} & \pmci{16.6}{6.0} & \pmci{17}{38} & \pmci{10.3}{2.2} \\
MyoHand Key Turn & \pmci{27}{76} & \pmci{36.1}{1.6} & \pmci{33}{143} & \pmci{4.6}{1.3} & \pmci{27}{76} & \pmci{36.1}{2.3} & \pmci{23}{63} & \pmci{34.2}{7.0} \\
MyoHand Obj Hold Hard & \pmci{50}{66} & \pmci{36.0}{1.4} & \pmci{70}{50} & \pmci{4.2}{.7} & \pmci{43}{29} & \pmci{29.7}{6.3} & \pmci{53}{63} & \pmci{16.5}{20.0} \\
MyoHand Obj Hold & $0$ & \pmci{35.2}{.5} & $0$ & \pmci{3.6}{.2} & $0$ & \pmci{34.6}{.7} & $0$ & \pmci{31.4}{6.3} \\
MyoHand Pen Twirl Hard & \pmci{47}{29} & \pmci{35.9}{.8} & \pmci{23}{14} & \pmci{3.3}{.6} & \pmci{50}{75} & \pmci{32.5}{5.2} & \pmci{27}{14} & \pmci{12.0}{5.9} \\
MyoHand Pen Twirl & \pmci{27}{57} & \pmci{35.1}{2.3} & \pmci{33}{143} & \pmci{3.1}{.5} & \pmci{27}{57} & \pmci{35.3}{1.0} & \pmci{27}{57} & \pmci{35.2}{3.0} \\
\bottomrule
\end{tabular}}
\end{center}
\end{table}

\begin{table}[h]
% \centering
\scriptsize
\setlength{\tabcolsep}{3.5pt}
\providecommand{\pmci}[2]{$#1_{\scriptscriptstyle \pm #2}$}
\caption{\textbf{ManiSkill2 per-task results.} 4 tasks, 3 seeds each. Mean and 95\% CI per task. Success rate (\%) and per-step latency (ms) for 4 key methods. \emph{all}: theoretical OOD mode.}
\label{tab:per_task_clean_maniskill2}
\begin{center}
\resizebox{\linewidth}{!}{%
\begin{tabular}{l r r r r r r r r}
\toprule
\textbf{Task} & \multicolumn{2}{c}{S2-alone (TD-MPC2)} & \multicolumn{2}{c}{S1 alone} & \multicolumn{2}{c}{Fast-TD-MPC (all, $\tau_{p50}$)} & \multicolumn{2}{c}{Fast-TD-MPC (all, $\tau_{p90}$)} \\
\cmidrule(lr){2-3} \cmidrule(lr){4-5} \cmidrule(lr){6-7} \cmidrule(lr){8-9}
 & \textbf{Succ.} & \textbf{Lat.}~{(ms)} & \textbf{Succ.} & \textbf{Lat.}~{(ms)} & \textbf{Succ.} & \textbf{Lat.}~{(ms)} & \textbf{Succ.} & \textbf{Lat.}~{(ms)} \\
\midrule
Lift Cube & $100$ & \pmci{46.1}{2.1} & \pmci{20}{50} & \pmci{8.2}{5.7} & \pmci{93}{29} & \pmci{38.1}{21.6} & \pmci{60}{25} & \pmci{17.0}{6.5} \\
Pick Cube & \pmci{87}{57} & \pmci{46.2}{2.3} & \pmci{53}{72} & \pmci{10.7}{1.3} & \pmci{73}{94} & \pmci{25.0}{8.1} & \pmci{50}{66} & \pmci{16.0}{1.3} \\
Pick Ycb & \pmci{27}{38} & \pmci{49.3}{3.8} & \pmci{3}{14} & \pmci{14.0}{2.4} & \pmci{27}{57} & \pmci{41.9}{8.0} & \pmci{13}{14} & \pmci{19.3}{9.1} \\
Stack Cube & \pmci{47}{115} & \pmci{41.9}{2.0} & $0$ & \pmci{9.7}{.6} & \pmci{23}{80} & \pmci{30.5}{11.9} & \pmci{7}{29} & \pmci{11.6}{2.0} \\
\bottomrule
\end{tabular}}
\end{center}
\end{table}

%% file: appendix/results_main_full_perturbed.tex
\section{Full Results under Disturbances}
\label{apdx:full_perturbation_results}

Tables~\ref{tab:appendix_perturbation_dmcontrol} -- \ref{tab:appendix_perturbation_maniskill2} present the full results under disturbances for each domain, including per-step latency, S2 invocation rate ($\rho$), and peak GPU memory, complementing Table~\ref{tab:perturbation_mild}. Tables~\ref{tab:per_task_perturbation_dmcontrol} -- \ref{tab:per_task_perturbation_maniskill2} provide corresponding per-task breakdowns.

\begin{table}[h]
% \centering
% \scriptsize
\small
\setlength{\tabcolsep}{4.5pt}
\providecommand{\pmci}[2]{$#1_{\scriptscriptstyle \pm #2}$}
\caption{\textbf{DMControl full results under disturbances.} 39 tasks, 3 seeds each. Mean and 95\% CI across all tasks. Return, per-step latency, S2 invocation rate ($\rho$), and peak GPU memory. \emph{all}: theoretical OOD mode; \emph{RG}: reward-gated mode.}
\label{tab:appendix_perturbation_dmcontrol}
\begin{center}
% \resizebox{\linewidth}{!}{%
\begin{tabular}{l r r r r}
\toprule
\textbf{Method} & \textbf{Return} & \textbf{Lat.}~{(ms)} & $\rho$~{(\%)} & \textbf{GPU}~{(MB)} \\
\midrule
S2-alone (TD-MPC2) & \pmci{616}{104} & \pmci{27.2}{1.1} & $100$ & \pmci{95}{0} \\
S2 policy-only & \pmci{390}{102} & \pmci{1.9}{.3} & $0$ & \pmci{66}{0} \\
S1 alone & \pmci{405}{98} & \pmci{1.7}{.3} & $0$ & \pmci{68}{0} \\
\midrule
Round-robin (S2 50\%) & \pmci{577}{102} & \pmci{14.5}{.7} & $50$ & \pmci{96}{0} \\
\midrule
Fast-TD-MPC (\emph{all}, $\tau_{p50}$) & \pmci{617}{105} & \pmci{24.5}{1.9} & \pmci{88}{5} & \pmci{91}{2} \\
Fast-TD-MPC (\emph{all}, $\tau_{p75}$) & \pmci{607}{103} & \pmci{19.3}{2.9} & \pmci{67}{10} & \pmci{86}{3} \\
Fast-TD-MPC (\emph{all}, $\tau_{p90}$) & \pmci{586}{101} & \pmci{13.6}{3.0} & \pmci{45}{11} & \pmci{81}{3} \\
Fast-TD-MPC (\emph{all}, $\tau_{p95}$) & \pmci{537}{102} & \pmci{9.8}{2.6} & \pmci{32}{10} & \pmci{77}{2} \\
Fast-TD-MPC (\emph{all}, $\tau_{p99}$) & \pmci{439}{101} & \pmci{5.6}{2.0} & \pmci{16}{8} & \pmci{72}{2} \\
\midrule
Fast-TD-MPC (\emph{RG}, $\tau_{p25}$) & \pmci{566}{104} & \pmci{17.5}{2.6} & \pmci{61}{10} & \pmci{85}{3} \\
Fast-TD-MPC (\emph{RG}, $\tau_{p50}$) & \pmci{543}{102} & \pmci{14.3}{2.5} & \pmci{49}{10} & \pmci{80}{3} \\
Fast-TD-MPC (\emph{RG}, $\tau_{p75}$) & \pmci{524}{102} & \pmci{12.2}{2.3} & \pmci{41}{9} & \pmci{80}{3} \\
\bottomrule
\end{tabular}
\end{center}
\end{table}

\begin{table}[h]
\centering
% \scriptsize
\small
\setlength{\tabcolsep}{4.5pt}
\providecommand{\pmci}[2]{$#1_{\scriptscriptstyle \pm #2}$}
\caption{\textbf{Meta-World full results under disturbances.} 50 tasks, 3 seeds each. Mean and 95\% CI across all tasks. Success rate, per-step latency, S2 invocation rate ($\rho$), and peak GPU memory. \emph{all}: theoretical OOD mode; \emph{RG}: reward-gated mode.}
\label{tab:appendix_perturbation_metaworld}
\begin{center}
% \resizebox{\linewidth}{!}{%
\begin{tabular}{l r r r r}
\toprule
\textbf{Method} & \textbf{Succ.} & \textbf{Lat.}~{(ms)} & $\rho$~{(\%)} & \textbf{GPU}~{(MB)} \\
\midrule
S2-alone (TD-MPC2) & \pmci{51}{10} & \pmci{26.8}{.2} & $100$ & \pmci{95}{0} \\
S2 policy-only & \pmci{49}{10} & \pmci{2.8}{.2} & $0$ & \pmci{66}{0} \\
S1 alone & \pmci{47}{10} & \pmci{2.6}{.2} & $0$ & \pmci{67}{0} \\
\midrule
Round-robin (S2 50\%) & \pmci{49}{11} & \pmci{14.7}{.2} & $50$ & \pmci{96}{0} \\
\midrule
Fast-TD-MPC (\emph{all}, $\tau_{p50}$) & \pmci{51}{11} & \pmci{27.2}{.2} & \pmci{99}{0} & \pmci{97}{0} \\
Fast-TD-MPC (\emph{all}, $\tau_{p75}$) & \pmci{51}{11} & \pmci{25.6}{.9} & \pmci{93}{3} & \pmci{96}{1} \\
Fast-TD-MPC (\emph{all}, $\tau_{p90}$) & \pmci{49}{10} & \pmci{22.8}{1.6} & \pmci{82}{6} & \pmci{95}{1} \\
Fast-TD-MPC (\emph{all}, $\tau_{p95}$) & \pmci{50}{10} & \pmci{21.8}{1.8} & \pmci{78}{7} & \pmci{93}{2} \\
Fast-TD-MPC (\emph{all}, $\tau_{p99}$) & \pmci{50}{10} & \pmci{20.2}{2.0} & \pmci{71}{8} & \pmci{92}{2} \\
\midrule
Fast-TD-MPC (\emph{RG}, $\tau_{p25}$) & \pmci{49}{11} & \pmci{20.9}{1.9} & \pmci{74}{7} & \pmci{91}{2} \\
Fast-TD-MPC (\emph{RG}, $\tau_{p50}$) & \pmci{50}{10} & \pmci{18.9}{2.1} & \pmci{66}{8} & \pmci{89}{2} \\
Fast-TD-MPC (\emph{RG}, $\tau_{p75}$) & \pmci{49}{10} & \pmci{17.4}{2.2} & \pmci{60}{8} & \pmci{87}{2} \\
\bottomrule
\end{tabular}
\end{center}
\end{table}

\begin{table}[h]
% \centering
% \scriptsize
\small
\setlength{\tabcolsep}{4.5pt}
\providecommand{\pmci}[2]{$#1_{\scriptscriptstyle \pm #2}$}
\caption{\textbf{MyoSuite full results under disturbances.} 10 tasks, 3 seeds each. Mean and 95\% CI across all tasks. Success rate, per-step latency, S2 invocation rate ($\rho$), and peak GPU memory. \emph{all}: theoretical OOD mode; \emph{RG}: reward-gated mode.}
\label{tab:appendix_perturbation_myosuite}
\begin{center}
% \resizebox{\linewidth}{!}{%
\begin{tabular}{l r r r r}
\toprule
\textbf{Method} & \textbf{Succ.} & \textbf{Lat.}~{(ms)} & $\rho$~{(\%)} & \textbf{GPU}~{(MB)} \\
\midrule
S2-alone (TD-MPC2) & \pmci{40}{22} & \pmci{35.7}{.5} & $100$ & \pmci{97}{0} \\
S2 policy-only & \pmci{17}{15} & \pmci{3.9}{.4} & $0$ & \pmci{67}{0} \\
S1 alone & \pmci{32}{15} & \pmci{3.5}{.3} & $0$ & \pmci{69}{0} \\
\midrule
Round-robin (S2 50\%) & \pmci{37}{20} & \pmci{19.4}{.3} & $50$ & \pmci{98}{0} \\
\midrule
Fast-TD-MPC (\emph{all}, $\tau_{p50}$) & \pmci{36}{20} & \pmci{32.2}{3.9} & \pmci{89}{13} & \pmci{94}{4} \\
Fast-TD-MPC (\emph{all}, $\tau_{p75}$) & \pmci{38}{22} & \pmci{27.2}{5.9} & \pmci{74}{19} & \pmci{89}{5} \\
Fast-TD-MPC (\emph{all}, $\tau_{p90}$) & \pmci{34}{22} & \pmci{22.9}{7.8} & \pmci{60}{24} & \pmci{87}{5} \\
Fast-TD-MPC (\emph{all}, $\tau_{p95}$) & \pmci{33}{20} & \pmci{20.0}{7.8} & \pmci{51}{25} & \pmci{81}{5} \\
Fast-TD-MPC (\emph{all}, $\tau_{p99}$) & \pmci{30}{18} & \pmci{13.3}{7.8} & \pmci{30}{25} & \pmci{79}{5} \\
\midrule
Fast-TD-MPC (\emph{RG}, $\tau_{p25}$) & \pmci{36}{22} & \pmci{25.7}{4.7} & \pmci{70}{15} & \pmci{92}{5} \\
Fast-TD-MPC (\emph{RG}, $\tau_{p50}$) & \pmci{33}{22} & \pmci{21.4}{5.6} & \pmci{56}{18} & \pmci{89}{5} \\
Fast-TD-MPC (\emph{RG}, $\tau_{p75}$) & \pmci{35}{24} & \pmci{17.7}{5.3} & \pmci{44}{17} & \pmci{82}{5} \\
\bottomrule
\end{tabular}
\end{center}
\end{table}

\begin{table}[h]
% \centering
% \scriptsize
\small
\setlength{\tabcolsep}{4.5pt}
\providecommand{\pmci}[2]{$#1_{\scriptscriptstyle \pm #2}$}
\caption{\textbf{ManiSkill2 full results under disturbances.} 3 tasks, 3 seeds each. Mean and 95\% CI across all tasks. Success rate, per-step latency, S2 invocation rate ($\rho$), and peak GPU memory. \emph{all}: theoretical OOD mode; \emph{RG}: reward-gated mode.}
\label{tab:appendix_perturbation_maniskill2}
% \resizebox{\linewidth}{!}{%
\begin{center}
\begin{tabular}{l r r r r}
\toprule
\textbf{Method} & \textbf{Succ.} & \textbf{Lat.}~{(ms)} & $\rho$~{(\%)} & \textbf{GPU}~{(MB)} \\
\midrule
S2-alone (TD-MPC2) & \pmci{66}{75} & \pmci{43.1}{4.7} & $100$ & \pmci{70}{0} \\
S2 policy-only & \pmci{63}{66} & \pmci{11.4}{5.7} & $0$ & \pmci{41}{0} \\
S1 alone & \pmci{17}{52} & \pmci{10.4}{4.2} & $0$ & \pmci{42}{0} \\
\midrule
Round-robin (S2 50\%) & \pmci{58}{89} & \pmci{27.1}{2.0} & $50$ & \pmci{72}{0} \\
\midrule
Fast-TD-MPC (\emph{all}, $\tau_{p50}$) & \pmci{66}{80} & \pmci{34.6}{18.0} & \pmci{74}{52} & \pmci{63}{10} \\
Fast-TD-MPC (\emph{all}, $\tau_{p75}$) & \pmci{49}{108} & \pmci{25.5}{22.3} & \pmci{44}{65} & \pmci{56}{10} \\
Fast-TD-MPC (\emph{all}, $\tau_{p90}$) & \pmci{46}{86} & \pmci{16.9}{10.5} & \pmci{18}{32} & \pmci{47}{6} \\
Fast-TD-MPC (\emph{all}, $\tau_{p95}$) & \pmci{32}{62} & \pmci{13.7}{5.2} & \pmci{10}{18} & \pmci{43}{0} \\
Fast-TD-MPC (\emph{all}, $\tau_{p99}$) & \pmci{19}{46} & \pmci{11.2}{3.1} & \pmci{2}{4} & \pmci{43}{0} \\
\midrule
Fast-TD-MPC (\emph{RG}, $\tau_{p25}$) & \pmci{62}{55} & \pmci{25.9}{25.6} & \pmci{46}{91} & \pmci{53}{10} \\
Fast-TD-MPC (\emph{RG}, $\tau_{p50}$) & \pmci{51}{61} & \pmci{23.1}{23.6} & \pmci{37}{84} & \pmci{53}{10} \\
Fast-TD-MPC (\emph{RG}, $\tau_{p75}$) & \pmci{43}{60} & \pmci{18.8}{14.7} & \pmci{23}{54} & \pmci{50}{8} \\
\bottomrule
\end{tabular}
\end{center}
\end{table}

\begin{table}[h]
% \centering
\scriptsize
\setlength{\tabcolsep}{3.5pt}
\providecommand{\pmci}[2]{$#1_{\scriptscriptstyle \pm #2}$}
\caption{\textbf{DMControl per-task results under disturbances.} 39 tasks, 3 seeds each. Mean and 95\% CI per task. Return and per-step latency (ms) for 4 key methods. \emph{all}: theoretical OOD mode.}
\label{tab:per_task_perturbation_dmcontrol}
\begin{center}
\resizebox{\linewidth}{!}{%
\begin{tabular}{l r r r r r r r r}
\toprule
\textbf{Task} & \multicolumn{2}{c}{S2-alone (TD-MPC2)} & \multicolumn{2}{c}{S1 alone} & \multicolumn{2}{c}{Fast-TD-MPC (all, $\tau_{p50}$)} & \multicolumn{2}{c}{Fast-TD-MPC (all, $\tau_{p90}$)} \\
\cmidrule(lr){2-3} \cmidrule(lr){4-5} \cmidrule(lr){6-7} \cmidrule(lr){8-9}
 & \textbf{Return} & \textbf{Lat.}~{(ms)} & \textbf{Return} & \textbf{Lat.}~{(ms)} & \textbf{Return} & \textbf{Lat.}~{(ms)} & \textbf{Return} & \textbf{Lat.}~{(ms)} \\
\midrule
Acrobot Swingup & \pmci{443}{44} & \pmci{25.4}{1.4} & \pmci{284}{171} & \pmci{1.3}{.0} & \pmci{386}{31} & \pmci{20.6}{13.7} & \pmci{368}{207} & \pmci{3.8}{1.5} \\
Cartpole Balance Sparse & \pmci{122}{132} & \pmci{25.6}{.4} & \pmci{118}{128} & \pmci{1.3}{.1} & \pmci{124}{135} & \pmci{25.6}{.5} & \pmci{123}{132} & \pmci{25.5}{1.2} \\
Cartpole Balance & \pmci{324}{245} & \pmci{25.8}{1.3} & \pmci{323}{117} & \pmci{1.5}{.0} & \pmci{322}{237} & \pmci{25.6}{.9} & \pmci{325}{242} & \pmci{26.1}{3.7} \\
Cartpole Swingup Sparse & \pmci{9}{19} & \pmci{25.3}{.5} & \pmci{4}{15} & \pmci{1.3}{.1} & \pmci{2}{4} & \pmci{21.4}{11.2} & \pmci{3}{3} & \pmci{11.9}{28.4} \\
Cartpole Swingup & \pmci{244}{151} & \pmci{25.8}{.4} & \pmci{213}{105} & \pmci{1.5}{.0} & \pmci{241}{168} & \pmci{24.5}{8.0} & \pmci{246}{155} & \pmci{18.6}{28.0} \\
Cheetah Jump & \pmci{796}{22} & \pmci{25.8}{.9} & \pmci{660}{250} & \pmci{1.3}{.0} & \pmci{786}{20} & \pmci{14.8}{2.4} & \pmci{787}{61} & \pmci{5.1}{.7} \\
\bottomrule
\end{tabular}}
\end{center}
\end{table}

\begin{table}[h]
% \centering
\scriptsize
\setlength{\tabcolsep}{3.5pt}
\providecommand{\pmci}[2]{$#1_{\scriptscriptstyle \pm #2}$}
\caption{\textbf{Meta-World per-task results under disturbances.} 50 tasks, 3 seeds each. Mean and 95\% CI per task. Success rate (\%) and per-step latency (ms) for 4 key methods. \emph{all}: theoretical OOD mode.}
\label{tab:per_task_perturbation_metaworld}
\begin{center}
\resizebox{\linewidth}{!}{%
\begin{tabular}{l r r r r r r r r}
\toprule
\textbf{Task} & \multicolumn{2}{c}{S2-alone (TD-MPC2)} & \multicolumn{2}{c}{S1 alone} & \multicolumn{2}{c}{Fast-TD-MPC (all, $\tau_{p50}$)} & \multicolumn{2}{c}{Fast-TD-MPC (all, $\tau_{p90}$)} \\
\cmidrule(lr){2-3} \cmidrule(lr){4-5} \cmidrule(lr){6-7} \cmidrule(lr){8-9}
 & \textbf{Succ.} & \textbf{Lat.}~{(ms)} & \textbf{Succ.} & \textbf{Lat.}~{(ms)} & \textbf{Succ.} & \textbf{Lat.}~{(ms)} & \textbf{Succ.} & \textbf{Lat.}~{(ms)} \\
\midrule
Assembly & \pmci{37}{29} & \pmci{26.6}{.3} & \pmci{23}{63} & \pmci{2.3}{.3} & \pmci{47}{52} & \pmci{27.9}{3.7} & \pmci{33}{38} & \pmci{26.2}{1.5} \\
Basketball & \pmci{13}{14} & \pmci{26.2}{1.2} & \pmci{20}{25} & \pmci{2.0}{.1} & \pmci{10}{25} & \pmci{27.4}{1.3} & \pmci{10}{25} & \pmci{26.9}{1.1} \\
Bin Picking & \pmci{7}{29} & \pmci{25.9}{.3} & \pmci{10}{25} & \pmci{2.1}{.1} & \pmci{7}{29} & \pmci{26.6}{1.1} & \pmci{3}{14} & \pmci{26.5}{1.0} \\
Box Close & \pmci{77}{14} & \pmci{27.4}{.6} & \pmci{53}{63} & \pmci{2.9}{.6} & \pmci{77}{29} & \pmci{28.5}{.9} & \pmci{77}{29} & \pmci{26.8}{2.6} \\
Button Press Topdown Wall & \pmci{97}{14} & \pmci{27.0}{.9} & $100$ & \pmci{2.8}{.2} & $100$ & \pmci{26.6}{1.5} & \pmci{93}{14} & \pmci{15.3}{5.6} \\
Button Press Topdown & $100$ & \pmci{26.9}{1.3} & $100$ & \pmci{2.7}{.2} & \pmci{97}{14} & \pmci{27.0}{1.3} & \pmci{97}{14} & \pmci{17.9}{7.4} \\
\bottomrule
\end{tabular}}
\end{center}
\end{table}

\begin{table}[h]
% \centering
\scriptsize
\setlength{\tabcolsep}{3.5pt}
\providecommand{\pmci}[2]{$#1_{\scriptscriptstyle \pm #2}$}
\caption{\textbf{MyoSuite per-task results under disturbances.} 10 tasks, 3 seeds each. Mean and 95\% CI per task. Success rate (\%) and per-step latency (ms) for 4 key methods. \emph{all}: theoretical OOD mode.}
\label{tab:per_task_perturbation_myosuite}
\begin{center}
\resizebox{\linewidth}{!}{%
\begin{tabular}{l r r r r r r r r}
\toprule
\textbf{Task} & \multicolumn{2}{c}{S2-alone (TD-MPC2)} & \multicolumn{2}{c}{S1 alone} & \multicolumn{2}{c}{Fast-TD-MPC (all, $\tau_{p50}$)} & \multicolumn{2}{c}{Fast-TD-MPC (all, $\tau_{p90}$)} \\
\cmidrule(lr){2-3} \cmidrule(lr){4-5} \cmidrule(lr){6-7} \cmidrule(lr){8-9}
 & \textbf{Succ.} & \textbf{Lat.}~{(ms)} & \textbf{Succ.} & \textbf{Lat.}~{(ms)} & \textbf{Succ.} & \textbf{Lat.}~{(ms)} & \textbf{Succ.} & \textbf{Lat.}~{(ms)} \\
\midrule
MyoHand Key Turn Hard & \pmci{30}{43} & \pmci{36.9}{1.4} & \pmci{17}{38} & \pmci{4.1}{.6} & \pmci{30}{75} & \pmci{15.2}{7.9} & \pmci{7}{29} & \pmci{10.5}{12.4} \\
MyoHand Key Turn & \pmci{40}{90} & \pmci{36.4}{1.6} & \pmci{47}{125} & \pmci{4.3}{.7} & \pmci{40}{90} & \pmci{37.0}{1.5} & \pmci{43}{103} & \pmci{34.8}{8.3} \\
MyoHand Obj Hold Hard & \pmci{70}{25} & \pmci{36.9}{2.5} & \pmci{50}{66} & \pmci{4.2}{.8} & \pmci{60}{25} & \pmci{33.8}{4.8} & \pmci{43}{29} & \pmci{16.7}{7.6} \\
MyoHand Obj Hold & $0$ & \pmci{36.0}{1.3} & $0$ & \pmci{3.5}{.3} & $0$ & \pmci{35.4}{1.5} & $0$ & \pmci{32.3}{12.0} \\
MyoHand Pen Twirl Hard & $30$ & \pmci{35.4}{1.9} & \pmci{27}{38} & \pmci{3.2}{.6} & \pmci{30}{50} & \pmci{32.7}{5.0} & \pmci{33}{38} & \pmci{13.9}{9.2} \\
MyoHand Pen Twirl & \pmci{3}{14} & \pmci{35.4}{1.2} & \pmci{13}{57} & \pmci{3.1}{.4} & \pmci{3}{14} & \pmci{35.5}{2.7} & \pmci{3}{14} & \pmci{35.7}{3.7} \\
\bottomrule
\end{tabular}}
\end{center}
\end{table}

\begin{table}[h]
% \centering
\scriptsize
\setlength{\tabcolsep}{3.5pt}
\providecommand{\pmci}[2]{$#1_{\scriptscriptstyle \pm #2}$}
\caption{\textbf{ManiSkill2 per-task results under disturbances.} 3 tasks, 3 seeds each. Mean and 95\% CI per task. Success rate (\%) and per-step latency (ms) for 4 key methods. \emph{all}: theoretical OOD mode.}
\label{tab:per_task_perturbation_maniskill2}
\begin{center}
\resizebox{\linewidth}{!}{%
\begin{tabular}{l r r r r r r r r}
\toprule
\textbf{Task} & \multicolumn{2}{c}{S2-alone (TD-MPC2)} & \multicolumn{2}{c}{S1 alone} & \multicolumn{2}{c}{Fast-TD-MPC (all, $\tau_{p50}$)} & \multicolumn{2}{c}{Fast-TD-MPC (all, $\tau_{p90}$)} \\
\cmidrule(lr){2-3} \cmidrule(lr){4-5} \cmidrule(lr){6-7} \cmidrule(lr){8-9}
 & \textbf{Succ.} & \textbf{Lat.}~{(ms)} & \textbf{Succ.} & \textbf{Lat.}~{(ms)} & \textbf{Succ.} & \textbf{Lat.}~{(ms)} & \textbf{Succ.} & \textbf{Lat.}~{(ms)} \\
\midrule
Lift Cube & \pmci{93}{14} & \pmci{44.3}{1.2} & \pmci{10}{25} & \pmci{8.5}{2.1} & \pmci{93}{14} & \pmci{42.7}{6.5} & \pmci{73}{29} & \pmci{21.8}{3.0} \\
Pick Cube & \pmci{70}{90} & \pmci{44.1}{1.5} & \pmci{40}{50} & \pmci{11.6}{2.5} & \pmci{73}{94} & \pmci{28.7}{9.3} & \pmci{57}{80} & \pmci{14.6}{2.9} \\
Stack Cube & \pmci{33}{100} & \pmci{40.9}{2.1} & $0$ & \pmci{11.2}{.7} & \pmci{30}{75} & \pmci{32.4}{10.4} & \pmci{7}{29} & \pmci{14.4}{6.5} \\
\bottomrule
\end{tabular}}
\end{center}
\end{table}

%% file: appendix/hardware_and_evaluation_setup.tex
\section{Evaluation Setup and Hardware Details}
\label{apdx:evaluation_setup_and_hardware}

\paragraph{Evaluation Metrics.}
We follow \citet{hansen2024tdmpc2}'s setup of training on 3 seeds and evaluating on the same 3 seeds, with 1 warmup + 10 evaluation episodes per task per seed.
We report return/success rate, per-step latency (ms), S2 invocation rate ($\rho$), and peak GPU memory.

\paragraph{Hardware.}
Per-step latency is measured with GPU synchronization at the start and end of each decision step (2 syncs), on a single CPU core with a dedicated empty GPU. DMControl, Meta-World, and MyoSuite run on an H100 node (2$\times$ Intel Xeon Gold 6430 CPUs, 2\,TB RAM, and 8$\times$ NVIDIA H100 80GB GPUs); ManiSkill2 on an A100 workstation (AMD EPYC 7V13 CPU, 216\,GB RAM, 1$\times$ A100 80GB).

%% file: appendix/mppi_warmstart_ablation.tex
\section{MPPI Planner Warm-Start Strategy Ablation}
\label{apdx:ablation_warmstart}

When the OOD gate triggers a System~1 to System~2 transition after $K$ timesteps, the planner's warm-start buffer (\texttt{\_prev\_mean}) contains a plan that is $K$ steps stale. We consider two initialization strategies:

\begin{itemize}[leftmargin=*,noitemsep]
    \item \textbf{Cold start:} Zero the buffer and discard all stale plans. The \gls{mppi} planner relies entirely on the policy prior $\pi_\theta$, which provides $N_{\pi}=24$ of $N=512$ candidate trajectories rolled out from the current latent through the world model.
    \item \textbf{$K$-shifted warm start:} Shift the buffer left by $K$ positions, salvaging the first $H{-}K$ future plans when $K < H{-}1$. If $K \geq H{-}1$, no valid plan survives and the strategy degenerates to cold start.
\end{itemize}

We evaluate both strategies across all 103 tasks from the 4 domains and 8 $\tau$ configurations (\emph{theoretical}: $p50$, $p75$, $p90$, $p95$, $p99$; \emph{reward-gated}: $p25$, $p50$, $p75$), yielding 824 paired comparisons on the clean benchmark and 816 under disturbance. A paired t-test finds no significant difference ($p > 0.05$) in return or S2 invocation rate between the two strategies on either benchmark, confirming that the policy prior provides sufficient fresh candidates on S1$\to$S2 transitions. We adopt cold start for its simplicity. Tables~\ref{tab:ablation_warmstart_clean} and~\ref{tab:ablation_warmstart_perturbation} report the full results.

\begin{table}[h]
\caption{\textbf{MPPI planner warm-start ablation across 4 domains (103 tasks, 3 seeds each).} Mean and 95\% CI across all tasks in each domain. \emph{all} denotes the \emph{theoretical} OOD mode; \emph{RG} denotes the \emph{reward-gated} mode. Cold start and $K$-shifted warm start yield equivalent results across all configurations.}
\label{tab:ablation_warmstart_clean}
%\centering
\scriptsize
\setlength{\tabcolsep}{4.5pt}
\providecommand{\pmci}[2]{$#1_{\scriptscriptstyle \pm #2}$}
\begin{center}
\begin{tabular}{l l *{8}{r}}
\toprule
 & & \multicolumn{2}{c}{\textbf{DMControl}} & \multicolumn{2}{c}{\textbf{Meta-World}} & \multicolumn{2}{c}{\textbf{MyoSuite}} & \multicolumn{2}{c}{\textbf{ManiSkill2}} \\
\cmidrule(lr){3-4} \cmidrule(lr){5-6} \cmidrule(lr){7-8} \cmidrule(lr){9-10}
\textbf{Config} & \textbf{Warm-Start} & \textbf{Return} & \textbf{Lat.}~{(ms)} & \textbf{Succ.}~{(\%)} & \textbf{Lat.}~{(ms)} & \textbf{Succ.}~{(\%)} & \textbf{Lat.}~{(ms)} & \textbf{Succ.}~{(\%)} & \textbf{Lat.}~{(ms)} \\
\midrule
\multirow{2}{*}{(\emph{all}, $\tau_{p50}$)} & Cold start & \pmci{829}{55} & \pmci{19.4}{2.4} & \pmci{85}{8} & \pmci{24.8}{1.1} & \pmci{39}{21} & \pmci{31.4}{3.6} & \pmci{54}{55} & \pmci{33.9}{12.1} \\
 & K-shifted & \pmci{829}{54} & \pmci{19.1}{2.5} & \pmci{85}{8} & \pmci{24.6}{1.1} & \pmci{38}{22} & \pmci{31.6}{3.5} & \pmci{55}{59} & \pmci{33.9}{13.9} \\
\midrule
\multirow{2}{*}{(\emph{all}, $\tau_{p75}$)} & Cold start & \pmci{823}{55} & \pmci{12.2}{2.7} & \pmci{85}{8} & \pmci{18.7}{1.5} & \pmci{37}{21} & \pmci{26.6}{5.8} & \pmci{47}{54} & \pmci{23.3}{10.7} \\
 & K-shifted & \pmci{821}{54} & \pmci{12.3}{2.7} & \pmci{85}{8} & \pmci{18.8}{1.5} & \pmci{39}{22} & \pmci{26.9}{5.8} & \pmci{48}{51} & \pmci{23.5}{11.6} \\
\midrule
\multirow{2}{*}{(\emph{all}, $\tau_{p90}$)} & Cold start & \pmci{785}{58} & \pmci{6.5}{2.0} & \pmci{85}{8} & \pmci{12.2}{1.6} & \pmci{37}{21} & \pmci{22.6}{7.6} & \pmci{32}{42} & \pmci{16.0}{5.2} \\
 & K-shifted & \pmci{785}{63} & \pmci{6.5}{2.0} & \pmci{85}{8} & \pmci{12.3}{1.6} & \pmci{35}{20} & \pmci{22.8}{7.2} & \pmci{32}{42} & \pmci{16.4}{5.8} \\
\midrule
\multirow{2}{*}{(\emph{all}, $\tau_{p95}$)} & Cold start & \pmci{745}{74} & \pmci{3.8}{1.4} & \pmci{85}{8} & \pmci{9.8}{1.5} & \pmci{38}{23} & \pmci{19.8}{7.6} & \pmci{28}{46} & \pmci{14.2}{3.9} \\
 & K-shifted & \pmci{744}{75} & \pmci{3.8}{1.3} & \pmci{86}{8} & \pmci{9.8}{1.5} & \pmci{39}{21} & \pmci{19.8}{7.3} & \pmci{28}{45} & \pmci{14.4}{4.8} \\
\midrule
\multirow{2}{*}{(\emph{all}, $\tau_{p99}$)} & Cold start & \pmci{650}{92} & \pmci{1.9}{.3} & \pmci{85}{8} & \pmci{6.2}{1.3} & \pmci{28}{17} & \pmci{10.9}{5.8} & \pmci{20}{39} & \pmci{11.5}{3.9} \\
 & K-shifted & \pmci{645}{93} & \pmci{1.8}{.3} & \pmci{86}{8} & \pmci{6.2}{1.3} & \pmci{31}{16} & \pmci{11.0}{6.0} & \pmci{20}{39} & \pmci{11.7}{4.4} \\
\midrule
\multirow{2}{*}{(\emph{RG}, $\tau_{p25}$)} & Cold start & \pmci{784}{68} & \pmci{10.8}{2.3} & \pmci{85}{8} & \pmci{12.3}{1.7} & \pmci{39}{22} & \pmci{24.5}{4.6} & \pmci{48}{43} & \pmci{28.5}{20.9} \\
 & K-shifted & \pmci{784}{67} & \pmci{10.8}{2.3} & \pmci{85}{8} & \pmci{12.4}{1.7} & \pmci{39}{21} & \pmci{24.4}{5.0} & \pmci{49}{39} & \pmci{29.4}{22.8} \\
\midrule
\multirow{2}{*}{(\emph{RG}, $\tau_{p50}$)} & Cold start & \pmci{764}{70} & \pmci{8.1}{1.8} & \pmci{85}{8} & \pmci{10.0}{1.5} & \pmci{35}{21} & \pmci{20.6}{5.2} & \pmci{43}{43} & \pmci{25.6}{18.9} \\
 & K-shifted & \pmci{768}{70} & \pmci{7.9}{1.8} & \pmci{85}{8} & \pmci{9.9}{1.5} & \pmci{41}{22} & \pmci{20.0}{5.3} & \pmci{46}{43} & \pmci{25.7}{18.5} \\
\midrule
\multirow{2}{*}{(\emph{RG}, $\tau_{p75}$)} & Cold start & \pmci{744}{74} & \pmci{6.4}{1.5} & \pmci{84}{8} & \pmci{8.6}{1.5} & \pmci{39}{21} & \pmci{15.7}{4.7} & \pmci{39}{51} & \pmci{20.9}{10.8} \\
 & K-shifted & \pmci{741}{74} & \pmci{6.4}{1.6} & \pmci{84}{8} & \pmci{8.6}{1.5} & \pmci{36}{20} & \pmci{16.8}{4.1} & \pmci{34}{48} & \pmci{19.9}{10.9} \\
\bottomrule
\end{tabular}
\end{center}
\end{table}

\begin{table}[h]
\caption{\textbf{MPPI planner warm-start ablation under disturbances across 4 domains (102 tasks, 3 seeds each).} Mean and 95\% CI across all tasks in each domain. \emph{all} denotes the \emph{theoretical} OOD mode; \emph{RG} denotes the \emph{reward-gated} mode. Cold start and $K$-shifted warm start yield equivalent results under perturbation.}
\label{tab:ablation_warmstart_perturbation}
%\centering
\scriptsize
\setlength{\tabcolsep}{4.5pt}
\providecommand{\pmci}[2]{$#1_{\scriptscriptstyle \pm #2}$}
\begin{center}
\begin{tabular}{l l *{8}{r}}
\toprule
 & & \multicolumn{2}{c}{\textbf{DMControl}} & \multicolumn{2}{c}{\textbf{Meta-World}} & \multicolumn{2}{c}{\textbf{MyoSuite}} & \multicolumn{2}{c}{\textbf{ManiSkill2}} \\
\cmidrule(lr){3-4} \cmidrule(lr){5-6} \cmidrule(lr){7-8} \cmidrule(lr){9-10}
\textbf{Config} & \textbf{Warm-Start} & \textbf{Return} & \textbf{Lat.}~{(ms)} & \textbf{Succ.}~{(\%)} & \textbf{Lat.}~{(ms)} & \textbf{Succ.}~{(\%)} & \textbf{Lat.}~{(ms)} & \textbf{Succ.}~{(\%)} & \textbf{Lat.}~{(ms)} \\
\midrule
\multirow{2}{*}{(\emph{all}, $\tau_{p50}$)} & Cold start & \pmci{617}{105} & \pmci{24.5}{1.9} & \pmci{51}{11} & \pmci{27.2}{.2} & \pmci{36}{20} & \pmci{32.2}{3.9} & \pmci{66}{80} & \pmci{34.6}{18.0} \\
 & K-shifted & \pmci{618}{105} & \pmci{24.5}{1.9} & \pmci{51}{11} & \pmci{27.1}{.2} & \pmci{38}{22} & \pmci{32.0}{3.2} & \pmci{69}{75} & \pmci{35.4}{17.7} \\
\midrule
\multirow{2}{*}{(\emph{all}, $\tau_{p75}$)} & Cold start & \pmci{607}{103} & \pmci{19.3}{2.9} & \pmci{51}{11} & \pmci{25.6}{.9} & \pmci{38}{22} & \pmci{27.2}{5.9} & \pmci{49}{108} & \pmci{25.5}{22.3} \\
 & K-shifted & \pmci{609}{103} & \pmci{19.4}{2.9} & \pmci{50}{10} & \pmci{25.4}{.9} & \pmci{40}{22} & \pmci{27.1}{6.1} & \pmci{54}{108} & \pmci{25.9}{21.4} \\
\midrule
\multirow{2}{*}{(\emph{all}, $\tau_{p90}$)} & Cold start & \pmci{586}{101} & \pmci{13.6}{3.0} & \pmci{49}{10} & \pmci{22.8}{1.6} & \pmci{34}{22} & \pmci{22.9}{7.8} & \pmci{46}{86} & \pmci{16.9}{10.5} \\
 & K-shifted & \pmci{584}{101} & \pmci{13.5}{3.0} & \pmci{50}{11} & \pmci{22.9}{1.6} & \pmci{33}{21} & \pmci{22.4}{8.0} & \pmci{43}{84} & \pmci{17.1}{10.0} \\
\midrule
\multirow{2}{*}{(\emph{all}, $\tau_{p95}$)} & Cold start & \pmci{537}{102} & \pmci{9.8}{2.6} & \pmci{50}{10} & \pmci{21.8}{1.8} & \pmci{33}{20} & \pmci{20.0}{7.8} & \pmci{32}{62} & \pmci{13.7}{5.2} \\
 & K-shifted & \pmci{529}{103} & \pmci{9.6}{2.5} & \pmci{51}{10} & \pmci{21.6}{1.8} & \pmci{32}{22} & \pmci{19.9}{7.8} & \pmci{34}{74} & \pmci{14.2}{5.7} \\
\midrule
\multirow{2}{*}{(\emph{all}, $\tau_{p99}$)} & Cold start & \pmci{439}{101} & \pmci{5.6}{2.0} & \pmci{50}{10} & \pmci{20.2}{2.0} & \pmci{30}{18} & \pmci{13.3}{7.8} & \pmci{19}{46} & \pmci{11.2}{3.1} \\
 & K-shifted & \pmci{446}{101} & \pmci{5.6}{1.9} & \pmci{50}{10} & \pmci{20.2}{2.0} & \pmci{31}{17} & \pmci{13.1}{7.8} & \pmci{20}{50} & \pmci{11.1}{2.7} \\
\midrule
\multirow{2}{*}{(\emph{RG}, $\tau_{p25}$)} & Cold start & \pmci{566}{104} & \pmci{17.5}{2.6} & \pmci{49}{11} & \pmci{20.9}{1.9} & \pmci{36}{22} & \pmci{25.7}{4.7} & \pmci{62}{55} & \pmci{25.9}{25.6} \\
 & K-shifted & \pmci{572}{103} & \pmci{17.3}{2.6} & \pmci{50}{11} & \pmci{20.8}{1.9} & \pmci{36}{22} & \pmci{25.2}{5.2} & \pmci{60}{52} & \pmci{25.6}{29.1} \\
\midrule
\multirow{2}{*}{(\emph{RG}, $\tau_{p50}$)} & Cold start & \pmci{543}{102} & \pmci{14.3}{2.5} & \pmci{50}{10} & \pmci{18.9}{2.1} & \pmci{33}{22} & \pmci{21.4}{5.6} & \pmci{51}{61} & \pmci{23.1}{23.6} \\
 & K-shifted & \pmci{546}{102} & \pmci{14.2}{2.5} & \pmci{50}{10} & \pmci{18.9}{2.1} & \pmci{37}{23} & \pmci{21.2}{5.9} & \pmci{59}{49} & \pmci{21.8}{23.5} \\
\midrule
\multirow{2}{*}{(\emph{RG}, $\tau_{p75}$)} & Cold start & \pmci{524}{102} & \pmci{12.2}{2.3} & \pmci{49}{10} & \pmci{17.4}{2.2} & \pmci{35}{24} & \pmci{17.7}{5.3} & \pmci{43}{60} & \pmci{18.8}{14.7} \\
 & K-shifted & \pmci{523}{102} & \pmci{12.2}{2.4} & \pmci{49}{10} & \pmci{17.4}{2.1} & \pmci{34}{23} & \pmci{18.0}{5.6} & \pmci{43}{60} & \pmci{19.8}{14.1} \\
\bottomrule
\end{tabular}
\end{center}
\end{table}

%% file: appendix/ood_gating_ablation.tex
\section{OOD Distance Metric Ablation}
\label{apdx:ablation_tanimoto}

We compare Mahalanobis distance against Tanimoto distance \citep{rogers1960tanimoto} as the OOD scoring function. Mahalanobis distance captures the full covariance structure of the in-distribution latents:
\begin{equation}
    g_{\text{maha}}(z_t) = (z_t - \bar{z})^\top \Sigma_{\text{reg}}^{-1} (z_t - \bar{z}),
\end{equation}
where $\bar{z} \in \mathbb{R}^{d_z}$ is the mean of the ID latents and $\Sigma_{\text{reg}} = \Sigma + \lambda \mathbf{I}$ is the $\ell_2$-regularized covariance (see Section~\ref{sec:method}), yielding an effective metric that accounts for inter-dimension correlations. Tanimoto distance measures angular and magnitude deviation from the mean without modeling correlations:
\begin{equation}
    g_{\text{tan}}(z_t) = 1 - \frac{z_t^\top \bar{z}}{\|z_t\|^2 + \|\bar{z}\|^2 - z_t^\top \bar{z}},
\end{equation}
where $z_t, \bar{z} \in \mathbb{R}^{d_z}$ and the denominator $\|z_t\|^2 + \|\bar{z}\|^2 - z_t^\top \bar{z}$ combines Euclidean and inner-product terms into a single similarity ratio. Tanimoto is computationally cheaper ($O(d_z)$ vs.\ $O(d_z^2)$ per step) and requires no covariance inversion during fitting. Both detectors are fitted on the same System~1 rollout trajectories and evaluated under identical $\tau$ percentiles.

Tables~\ref{tab:ablation_tanimoto_clean} and~\ref{tab:ablation_tanimoto_perturbation} report the full results. Under nominal conditions, a paired t-test across all task-config comparisons finds no significant difference in return ($t{=}0.85$, $p{=}0.40$), although Mahalanobis achieves substantially higher success on Meta-World (85\% vs.\ 70\% at $\tau_{p50}$). In the \emph{theoretical} mode, Tanimoto yields equal or lower latency at every $\tau$ percentile, reflecting fewer S2 invocations; in the \emph{reward-gated} mode, it is often slower. Under perturbation, Mahalanobis yields significantly higher return ($t{=}2.33$, $p{=}0.02$) by invoking S2 more often. We therefore adopt Mahalanobis for its stronger robustness.

\begin{table}[h]
\caption{\textbf{OOD metric ablation across 4 domains (103 tasks, 3 seeds each).} Mean and 95\% CI across all tasks in each domain. \emph{all} denotes the \emph{theoretical} OOD mode; \emph{RG} denotes the \emph{reward-gated} mode.}
\label{tab:ablation_tanimoto_clean}
\scriptsize
\setlength{\tabcolsep}{4.5pt}
\providecommand{\pmci}[2]{$#1_{\scriptscriptstyle \pm #2}$}
\begin{center}
\begin{tabular}{l l *{8}{r}}
\toprule
 & & \multicolumn{2}{c}{\textbf{DMControl}} & \multicolumn{2}{c}{\textbf{Meta-World}} & \multicolumn{2}{c}{\textbf{MyoSuite}} & \multicolumn{2}{c}{\textbf{ManiSkill2}} \\
\cmidrule(lr){3-4} \cmidrule(lr){5-6} \cmidrule(lr){7-8} \cmidrule(lr){9-10}
\textbf{Config} & \textbf{Metric} & \textbf{Return} & \textbf{Lat.}~{(ms)} & \textbf{Succ.}~{(\%)} & \textbf{Lat.}~{(ms)} & \textbf{Succ.}~{(\%)} & \textbf{Lat.}~{(ms)} & \textbf{Succ.}~{(\%)} & \textbf{Lat.}~{(ms)} \\
\midrule
\multirow{2}{*}{(\emph{all}, $\tau_{p50}$)} & Maha. & \pmci{829}{55} & \pmci{19.4}{2.4} & \pmci{85}{8} & \pmci{24.8}{1.1} & \pmci{39}{21} & \pmci{31.4}{3.6} & \pmci{54}{55} & \pmci{33.9}{12.1} \\
 & Tan. & \pmci{819.9}{53.5} & \pmci{15.4}{2.4} & \pmci{69.9}{6.3} & \pmci{21.5}{1.2} & \pmci{40.2}{21} & \pmci{27.8}{4.5} & \pmci{48.0}{33.3} & \pmci{24.7}{6.7} \\
\midrule
\multirow{2}{*}{(\emph{all}, $\tau_{p75}$)} & Maha. & \pmci{823}{55} & \pmci{12.2}{2.7} & \pmci{85}{8} & \pmci{18.7}{1.5} & \pmci{37}{21} & \pmci{26.6}{5.8} & \pmci{47}{54} & \pmci{23.3}{10.7} \\
 & Tan. & \pmci{796.9}{56.1} & \pmci{9.9}{2.5} & \pmci{69.4}{6.3} & \pmci{13.6}{1.2} & \pmci{38.4}{21.2} & \pmci{19.1}{5.5} & \pmci{41.1}{34.8} & \pmci{18.8}{7.3} \\
\midrule
\multirow{2}{*}{(\emph{all}, $\tau_{p90}$)} & Maha. & \pmci{785}{58} & \pmci{6.5}{2.0} & \pmci{85}{8} & \pmci{12.2}{1.6} & \pmci{37}{21} & \pmci{22.6}{7.6} & \pmci{32}{42} & \pmci{16.0}{5.2} \\
 & Tan. & \pmci{733.8}{68.9} & \pmci{5.2}{1.8} & \pmci{69.4}{6.3} & \pmci{6.3}{0.8} & \pmci{38.6}{20} & \pmci{10.6}{3.8} & \pmci{28.3}{28.3} & \pmci{12.3}{1.6} \\
\midrule
\multirow{2}{*}{(\emph{all}, $\tau_{p95}$)} & Maha. & \pmci{745}{74} & \pmci{3.8}{1.4} & \pmci{85}{8} & \pmci{9.8}{1.5} & \pmci{38}{23} & \pmci{19.8}{7.6} & \pmci{28}{46} & \pmci{14.2}{3.9} \\
 & Tan. & \pmci{696.6}{79.6} & \pmci{3.5}{0.9} & \pmci{69.9}{6.4} & \pmci{4.6}{0.6} & \pmci{33.7}{19.6} & \pmci{7.4}{2.1} & \pmci{23.3}{24.8} & \pmci{11.4}{1.8} \\
\midrule
\multirow{2}{*}{(\emph{all}, $\tau_{p99}$)} & Maha. & \pmci{650}{92} & \pmci{1.9}{.3} & \pmci{85}{8} & \pmci{6.2}{1.3} & \pmci{28}{17} & \pmci{10.9}{5.8} & \pmci{20}{39} & \pmci{11.5}{3.9} \\
 & Tan. & \pmci{650.9}{91.2} & \pmci{1.9}{0.2} & \pmci{69.1}{6.8} & \pmci{3.2}{0.3} & \pmci{32.4}{18.7} & \pmci{4.5}{0.7} & \pmci{19.5}{21.2} & \pmci{10.7}{2.1} \\
\midrule
\multirow{2}{*}{(\emph{RG}, $\tau_{p25}$)} & Maha. & \pmci{784}{68} & \pmci{10.8}{2.3} & \pmci{85}{8} & \pmci{12.3}{1.7} & \pmci{39}{22} & \pmci{24.5}{4.6} & \pmci{48}{43} & \pmci{28.5}{20.9} \\
 & Tan. & \pmci{807.1}{58} & \pmci{12.0}{2.1} & \pmci{69.9}{6.6} & \pmci{12.3}{1.6} & \pmci{41.1}{21.5} & \pmci{24.3}{4.5} & \pmci{46.1}{34.5} & \pmci{25.0}{11.3} \\
\midrule
\multirow{2}{*}{(\emph{RG}, $\tau_{p50}$)} & Maha. & \pmci{764}{70} & \pmci{8.1}{1.8} & \pmci{85}{8} & \pmci{10.0}{1.5} & \pmci{35}{21} & \pmci{20.6}{5.2} & \pmci{43}{43} & \pmci{25.6}{18.9} \\
 & Tan. & \pmci{786.7}{63.5} & \pmci{8.6}{1.7} & \pmci{69.3}{6.7} & \pmci{11.4}{1.5} & \pmci{40.6}{21.2} & \pmci{19.8}{4.6} & \pmci{43.4}{29.7} & \pmci{23.4}{10.3} \\
\midrule
\multirow{2}{*}{(\emph{RG}, $\tau_{p75}$)} & Maha. & \pmci{744}{74} & \pmci{6.4}{1.5} & \pmci{84}{8} & \pmci{8.6}{1.5} & \pmci{39}{21} & \pmci{15.7}{4.7} & \pmci{39}{51} & \pmci{20.9}{10.8} \\
 & Tan. & \pmci{769.1}{70.1} & \pmci{6.6}{1.3} & \pmci{69.2}{6.7} & \pmci{9.6}{1.3} & \pmci{38.1}{19.3} & \pmci{17.4}{4.2} & \pmci{38.0}{30.7} & \pmci{19.1}{7.4} \\
\bottomrule
\end{tabular}
\end{center}
\end{table}

\begin{table}[h]
\caption{\textbf{OOD metric ablation under disturbances across 4 domains (102 tasks, 3 seeds each).} Mean and 95\% CI across all tasks in each domain. \emph{all} denotes the \emph{theoretical} OOD mode; \emph{RG} denotes the \emph{reward-gated} mode.}
\label{tab:ablation_tanimoto_perturbation}
%\centering
\scriptsize
\setlength{\tabcolsep}{4.5pt}
\providecommand{\pmci}[2]{$#1_{\scriptscriptstyle \pm #2}$}
\begin{center}
\begin{tabular}{l l *{8}{r}}
\toprule
 & & \multicolumn{2}{c}{\textbf{DMControl}} & \multicolumn{2}{c}{\textbf{Meta-World}} & \multicolumn{2}{c}{\textbf{MyoSuite}} & \multicolumn{2}{c}{\textbf{ManiSkill2}} \\
\cmidrule(lr){3-4} \cmidrule(lr){5-6} \cmidrule(lr){7-8} \cmidrule(lr){9-10}
\textbf{Config} & \textbf{Metric} & \textbf{Return} & \textbf{Lat.}~{(ms)} & \textbf{Succ.}~{(\%)} & \textbf{Lat.}~{(ms)} & \textbf{Succ.}~{(\%)} & \textbf{Lat.}~{(ms)} & \textbf{Succ.}~{(\%)} & \textbf{Lat.}~{(ms)} \\
\midrule
\multirow{2}{*}{(\emph{all}, $\tau_{p50}$)} & Maha. & \pmci{617}{105} & \pmci{24.5}{1.9} & \pmci{51}{11} & \pmci{27.2}{.2} & \pmci{36}{20} & \pmci{32.2}{3.9} & \pmci{66}{80} & \pmci{34.6}{18.0} \\
 & Tan. & \pmci{606.0}{103.5} & \pmci{20.5}{2.2} & \pmci{42.7}{8.4} & \pmci{25.9}{0.8} & \pmci{34.8}{20.8} & \pmci{27.3}{5.2} & \pmci{57.9}{32.6} & \pmci{26.7}{8.1} \\
\midrule
\multirow{2}{*}{(\emph{all}, $\tau_{p75}$)} & Maha. & \pmci{607}{103} & \pmci{19.3}{2.9} & \pmci{51}{11} & \pmci{25.6}{.9} & \pmci{38}{22} & \pmci{27.2}{5.9} & \pmci{49}{108} & \pmci{25.5}{22.3} \\
 & Tan. & \pmci{571.4}{99.5} & \pmci{14.8}{2.7} & \pmci{43.0}{8.4} & \pmci{22.0}{1.7} & \pmci{35.0}{20.4} & \pmci{19.6}{6} & \pmci{45}{38.6} & \pmci{19.3}{10.8} \\
\midrule
\multirow{2}{*}{(\emph{all}, $\tau_{p90}$)} & Maha. & \pmci{586}{101} & \pmci{13.6}{3.0} & \pmci{49}{10} & \pmci{22.8}{1.6} & \pmci{34}{22} & \pmci{22.9}{7.8} & \pmci{46}{86} & \pmci{16.9}{10.5} \\
 & Tan. & \pmci{503.0}{97.8} & \pmci{10.1}{2.6} & \pmci{42.1}{8.4} & \pmci{14.5}{2.3} & \pmci{34.3}{20.4} & \pmci{12.6}{4.2} & \pmci{31.4}{34} & \pmci{13.9}{3.5} \\
\midrule
\multirow{2}{*}{(\emph{all}, $\tau_{p95}$)} & Maha. & \pmci{537}{102} & \pmci{9.8}{2.6} & \pmci{50}{10} & \pmci{21.8}{1.8} & \pmci{33}{20} & \pmci{20.0}{7.8} & \pmci{32}{62} & \pmci{13.7}{5.2} \\
 & Tan. & \pmci{461.4}{99.9} & \pmci{7.8}{2.4} & \pmci{41.9}{8.3} & \pmci{12.5}{2.3} & \pmci{33.2}{20.1} & \pmci{8.2}{2.9} & \pmci{27.8}{27.8} & \pmci{11.5}{0.9} \\
\midrule
\multirow{2}{*}{(\emph{all}, $\tau_{p99}$)} & Maha. & \pmci{439}{101} & \pmci{5.6}{2.0} & \pmci{50}{10} & \pmci{20.2}{2.0} & \pmci{30}{18} & \pmci{13.3}{7.8} & \pmci{19}{46} & \pmci{11.2}{3.1} \\
 & Tan. & \pmci{413.3}{99.9} & \pmci{5.1}{2.1} & \pmci{41.6}{8.1} & \pmci{10.8}{2.2} & \pmci{32.7}{19.4} & \pmci{5.4}{1.8} & \pmci{21.5}{23.6} & \pmci{10.0}{1.1} \\
\midrule
\multirow{2}{*}{(\emph{RG}, $\tau_{p25}$)} & Maha. & \pmci{566}{104} & \pmci{17.5}{2.6} & \pmci{49}{11} & \pmci{20.9}{1.9} & \pmci{36}{22} & \pmci{25.7}{4.7} & \pmci{62}{55} & \pmci{25.9}{25.6} \\
 & Tan. & \pmci{585.2}{102.4} & \pmci{17.6}{2.3} & \pmci{42.5}{8.4} & \pmci{19.2}{1.9} & \pmci{40.6}{20} & \pmci{24.1}{4.8} & \pmci{54.3}{32.9} & \pmci{21.3}{7} \\
\midrule
\multirow{2}{*}{(\emph{RG}, $\tau_{p50}$)} & Maha. & \pmci{543}{102} & \pmci{14.3}{2.5} & \pmci{50}{10} & \pmci{18.9}{2.1} & \pmci{33}{22} & \pmci{21.4}{5.6} & \pmci{51}{61} & \pmci{23.1}{23.6} \\
 & Tan. & \pmci{562.5}{102.1} & \pmci{14.2}{2.4} & \pmci{42.7}{8.5} & \pmci{18.1}{2} & \pmci{38.6}{20.5} & \pmci{21.2}{4.6} & \pmci{53.5}{36.7} & \pmci{20.5}{9} \\
\midrule
\multirow{2}{*}{(\emph{RG}, $\tau_{p75}$)} & Maha. & \pmci{524}{102} & \pmci{12.2}{2.3} & \pmci{49}{10} & \pmci{17.4}{2.2} & \pmci{35}{24} & \pmci{17.7}{5.3} & \pmci{43}{60} & \pmci{18.8}{14.7} \\
 & Tan. & \pmci{539.3}{103.2} & \pmci{12.2}{2.4} & \pmci{42.7}{8.5} & \pmci{16.7}{2.1} & \pmci{37.6}{19.4} & \pmci{18.5}{4.1} & \pmci{45.4}{31.7} & \pmci{15.2}{2.3} \\
\bottomrule
\end{tabular}
\end{center}
\end{table}

%% file: appendix/implementation_details.tex
\section{Implementation Details}
\label{apdx:implementation_details}

\textbf{Single configuration.} All Fast-TD-MPC components (architecture, training, and OOD detection) use the same hyperparameters across all 103 tasks. No per-task tuning is performed; $\tau$ is set automatically for each task-seed from
held-out OOD scores using the same rule for all tasks (the median for \emph{theoretical} mode; the midpoint of the ID and OOD medians for \emph{reward-gated} mode; see Section~\ref{sec:method}).

\subsection{System 1: Amortized Policy Architecture}

\paragraph{Model Architecture.} All components of TD-MPC2 are parameterized by MLPs with LayerNorm and Mish activations. The encoder $\mathcal{E}_{\theta}$ and dynamics model $\mathcal{D}_{\theta}$ additionally use Simplicial Normalization (SimNorm) \citep{lavoie2022simnorm} on their final layer to normalize the latent representation $z$, biasing it towards sparsity without enforcing hard discrete constraints. Mirroring this architecture, the amortized policy $\pi_{\text{S1}}$ is a 4-layer MLP: three intermediate layers of linear $\to$ LayerNorm $\to$ Mish, followed by a final SimNorm layer. The backbone takes the frozen latent $z \in \mathbb{R}^{d_z}$ ($d_z = 512$) as input, maps to a 256-dimensional hidden representation, and a deterministic linear head produces the action. We use history length $L = 1$ and chunk size $C = 1$ (single-step prediction). Following TD-MPC2, actions are squashed with $\tanh$ at the output and dropout (rate 0.1) is applied to the first layer only. We summarize the $\pi_{\text{S1}}$ architecture using PyTorch-like notation:

\begin{verbatim}
backbone = SimNormMLPBackbone(
    (0): NormedLinear(512, 256, act=Mish, dropout=0.1)
    (1): NormedLinear(256, 256, act=Mish)
    (2): NormedLinear(256, 256, act=Mish)
    (3): NormedLinear(256, 256, act=SimNorm(dim=8))
)
head = DeterministicHead(
    Linear(256, action_dim)
)
\end{verbatim}

\textbf{Parameter count.} The backbone contains 330{,}752 parameters; the head contains $256 \times d_a + d_a$ parameters. Total: ${\sim}$332K for $d_a = 6$ (DMControl cheetah) to ${\sim}$341K for $d_a = 39$ (MyoSuite). Compared to TD-MPC2's 7.9M parameters, this represents a $23\times$ reduction.

\subsection{System 1 Training}

\textbf{Expert demonstrations.} Expert demonstrations $\mathcal{D}_{\text{exp}} = \{(z_t, a^{*}_t)\}$ are collected by rolling out the pre-trained TD-MPC2 planner (3 seeds per task, 500 episodes per seed) and recording the latent $z_t = \mathcal{E}_\theta(s_t)$ and the planner's executed action $a^{*}_t$. Episodes are split 8:1:1 into training, validation, and testing.

\textbf{Loss.} We use L1 loss: $\ell(\hat{a}, a^{*}) = |\hat{a} - a^{*}|$, where $|\hat{a} - a^{*}|$ is the element-wise absolute difference, summed over the action dimension $d_a$.

\textbf{Optimizer.} AdamW \citep{loshchilov2019decoupled} with learning rate (lr) $5 \times 10^{-5}$, weight decay $10^{-4}$, betas $(0.9, 0.999)$, cosine learning rate schedule with 10\% warmup and minimum lr $10^{-6}$.

\textbf{Training.} Batch size 512, gradient clipping at 1.0, maximum 200 epochs with early stopping (patience 15, minimum delta $10^{-4}$). All models are trained to full convergence.

\subsection{OOD Detector}

The OOD detector uses Mahalanobis distance in the frozen encoder's latent space, operating on $d_z = 512$-dimensional latent features ($z_t$ with history length $L = 1$). The in-distribution mean $\bar{z}$ and full covariance $\Sigma$ are computed from 400 System~1 rollout episodes per task-seed. We apply $\ell_2$ shrinkage: $\Sigma_{\text{reg}} = \Sigma + \lambda \mathbf{I}$ with $\lambda = 0.01$. The scoring function is $g(z_t) = (z_t - \bar{z})^\top \Sigma_{\text{reg}}^{-1} (z_t - \bar{z})$. The threshold $\tau$ is computed on a separate held-out set of 100 rollout episodes.

We consider two labeling modes for constructing the ID set:
\begin{itemize}[leftmargin=*]
    \item \textbf{Theoretical} (\emph{all}): all visited latents from System~1 rollouts are treated as in-distribution. The threshold $\tau$ defaults to the median of $g(z_t)$ scores on the held-out set.
    \item \textbf{Reward-gated} (\emph{RG}): a state is labeled in-distribution only if its total per-step reward or task completion proxy obtained from the environment meets or exceeds the TD-MPC2 expert's per-step counterpart at the same timestep. The threshold $\tau$ defaults to the midpoint between the median ID and OOD scores.
\end{itemize}

\subsection{System 2}

\textbf{System 2.} The \gls{mppi} planner (System~2) is used as-is from TD-MPC2, with default hyperparameters: planning horizon $H = 3$, 6 refinement iterations, $N=512$ samples, $N_\pi = 24$ policy-prior trajectories, top-$k=64$ elite trajectories, min/std $0.05$/$2.0$, temperature $0.5$.

%% file: appendix/extended_related_work.tex
\section{Extended Related Work}
\label{apdx:related_work}

\paragraph{Data-Driven MPC and the TD-MPC Family.}
Data-driven model predictive control combines learned world models with online trajectory optimization for planning or data generation.
TD-MPC2 \citep{hansen2024tdmpc2} combines a task-oriented latent dynamics model with \gls{mppi} planning and \gls{td}-learning, achieving state-of-the-art performance across 104 continuous control tasks with a single set of hyperparameters, outperforming strong model-free baselines such as SAC \citep{haarnoja2018sac} and model-based methods such as DreamerV3 \citep{hafner2023dreamerv3}.
Its growing ecosystem addresses training efficiency \citep{evers2026efficienttdmpc}, humanoid locomotion \citep{nguyen2025tdgrpc,nguyen2025doublyaware}, hierarchical planning \citep{chitnis2023iql}, policy constraints \citep{lin2025tdmpc2}, discrete latent representations \citep{scannell2025dcmwm}, and world model reliability \citep{hansen2026hallucination}.
Concurrent approaches include gradient-based planners that replace sampling-based methods with differentiable optimization \citep{dreammpc2026}.
Fast-TD-MPC builds directly on the TD-MPC2 framework but, unlike these works, targets \emph{inference-time} acceleration rather than training improvements or architectural modifications.

\paragraph{Planner Amortization and Speedup.}
A natural approach to reducing the cost of online planning is to distill the planner into a lightweight feed-forward policy.
\citet{kuzmenko2025tdmpc_opt} distill a 317M-parameter multi-task TD-MPC2 agent into a smaller quantized model, but \gls{mppi} still runs at every step.
\citet{wang2025bmpc} learn a policy by imitating an \gls{mpc} expert via lazy reanalyze, but similarly still execute the planner at deployment, with the policy only seeding the optimization.
\citet{nguyen2026latentgeometry} replace iterative planning entirely with a goal-conditioned inverse dynamics model, achieving 100-130$\times$ cost reduction, but provide no fallback when the amortized policy fails on out-of-distribution states.
\citet{gao2026fastlewm} accelerate the planner itself by replacing autoregressive latent rollouts with parallel action-prefix prediction, and \citet{huang2026leflow} amortize iterative planning with a generative flow-matching model; both still execute the full planning loop.
\citet{brudermuller2025gpc} similarly amortize sampling-based \gls{mpc} via flow matching but always run refinement iterations.
Alternative acceleration mechanisms include GPU-parallelized \gls{mpc} solvers \citep{bravopalacios2026turboempc} and gradient-based \gls{mpc} in latent space \citep{dreammpc2026}, which are orthogonal to our approach.
Fast-TD-MPC differs by training a \emph{separately amortized} policy that can entirely replace the planner on in-distribution states, while retaining the full \gls{mppi} planner as a fallback triggered by OOD detection.

\paragraph{Adaptive and Intermittent Replanning.}
The idea of conditionally skipping replanning has a long history in classical control theory.
\citet{tabuada2007event} introduced event-triggered control, showing that a stabilizing controller need not execute at every timestep but only when a state-dependent condition is violated.
\citet{heemels2012event} provided a canonical survey formalizing event-triggered and self-triggered control.
\citet{hashimoto2017event} proposed \emph{intermittent} \gls{mpc} for nonlinear systems, deliberately skipping replanning when the system is tracking well. This is the closest classical analogue to Fast-TD-MPC's conditional use of the fast policy.

In the data-driven \gls{mpc} setting, \citet{cheng2026adarep} propose AdaReP, a training-free wrapper that adapts replanning tolerance based on model mismatch, achieving over 80\% fewer planner queries.
However, AdaReP reuses the \emph{last} cached plan rather than executing a separately learned policy, and its gate is a continuous mismatch threshold rather than a learned OOD detector.
\citet{lin2025speculation} apply speculative execution --- a paradigm originating in LLM inference \citep{leviathan2023speculative,chen2023speculative}, directly to TD-MPC2, generating action queues from the planner's own latent rollouts and falling back to full replanning on mismatch.
This approach is comparable to Fast-TD-MPC, but differs in three key aspects: (i) the fast path uses the planner's own predictions rather than a separately trained amortized policy, (ii) the gate is a mismatch threshold rather than a learned OOD detector, and (iii) their fast path executes open-loop action queues, whereas Fast-TD-MPC's System~1 acts closed-loop with per-step OOD gating.

\paragraph{Dual-Process Control and Adaptive Computation.}
The dual-process framework of distinguishing a fast, reflexive System 1 from a slow, deliberative System 2 originates in cognitive science \citep{kahneman2011thinking,daw2005uncertainty,botvinick2009planning}.
A growing body of work adopts this framing for control: \citet{labiosa2026whentoplan} study learning when to switch between reactive control and deliberative planning, while \citet{li2026elastic} learn state-dependent compute schedules for generative control policies.
\citet{muppidi2026findingtime} learn planning budgets in real-time RL where the environment does not wait for the agent, directly addressing the think-vs-act trade-off that motivates Fast-TD-MPC.
Within the adaptive computation paradigm, \citet{chun2025dasip} dynamically adjust integration steps for stochastic interpolant policies based on task difficulty, and \citet{salla2026pogp} learn a prefix-optimal stopping criterion for diffusion policies.
\citet{zhao2026gatedgeobon} propose a consistency gate that invokes expensive Best-of-$N$ selection only when the initial rollout appears internally inconsistent, recovering most of the benefit at a fraction of the cost: a design philosophy shared with Fast-TD-MPC.
Concurrent work has proliferated in the VLA domain, including speculation-verification frameworks \citep{svvla2026}, uncertainty-driven ``cognitive clutch'' mechanisms \citep{li2026vlaattc}, and learned binary gates for cheap/expensive policy switching \citep{aegis2026,arms2026}.
\citet{yan2026dualprocess} apply the dual-process framework to motion planning with symbolic solvers as System 2.
However, these works operate in VLA, symbolic planning, or diffusion policy domains. Fast-TD-MPC is, to our knowledge, the first to leverage the dual-process framework within MBRL, where the learned world model and \gls{mppi} planner provide a principled System 2 that other domains approximate with less grounded alternatives.

\paragraph{Out-of-Distribution Detection in RL.}
OOD detection is a mature technique in deep learning \citep{hendrycks2017baseline,lee2018mahalanobis}, but its application to RL and control settings has received comparatively less attention.
\citet{zhang2021mdx} introduce a distance-based anomaly detection framework for deep RL, directly comparable to our Mahalanobis approach.
\citet{kaur2023codit} apply conformal OOD detection to time-series data for safety-critical autonomous systems, and \citet{tang2026autointervene} use OOD detection to trigger intervention in action-chunking policies --- a gating mechanism analogous to Fast-TD-MPC's, though in the VLA domain.
Other works provide formal guarantees for OOD detection in deep RL via transition estimation \citep{prashant2025guaranteeing} and benchmark OOD detection methods specifically for RL \citep{mittag2026oodrlbench}.
Fast-TD-MPC repurposes Mahalanobis OOD detection \citep{lee2018mahalanobis} as a per-timestep routing mechanism between an amortized policy and a model-based planner --- using distributional shift not for anomaly alerting, but as a signal for adaptive compute allocation, a role that, to our knowledge, has not been explored.